# Evaluating Machine Translation Datasets for Low-Web Data Languages: A Gendered Lens

**Hellina Hailu Nigatu**[1]*, **Bethelhem Yemane Mamo**[2], **Bontu Fufa Balcha**[2], **Debora Taye Tesfaye**[2], **Elbethel Daniel Zewdie**[2], **Ikram Behiru Nesiru**[2], **Jitu Ewnetu Hailu**[2], **Senait Mengesha Yayo**[3]

[1]UC Berkeley, [2]Addis Ababa University, [3]Addis Ababa Science and Technology University,

**Correspondence:** hellina_nigatu@berkeley.edu

## Abstract

As low-resourced languages are increasingly incorporated into NLP research, there is an emphasis on collecting large-scale datasets. But in prioritizing quantity over quality, we risk 1) building language technologies that perform poorly for these languages and 2) producing harmful content that perpetuates societal biases. In this paper, we investigate the quality of Machine Translation (MT) datasets for three low-resourced languages–Afan Oromo, Amharic, and Tigrinya, with a focus on the gender representation in the datasets. Our findings demonstrate that while training data has a large representation of political and religious domain text, benchmark datasets are focused on news, health, and sports. We also found a large skew towards the male gender–in names of persons, the grammatical gender of verbs, and in stereotypical depictions in the datasets. Further, we found harmful and toxic depictions against women, which were more prominent for the language with the largest amount of data, underscoring that quantity does not guarantee quality. We hope that our work inspires further inquiry into the datasets collected for low-resourced languages and prompts early mitigation of harmful content[1]. WARNING: This paper contains discussion of NSFW content that some may find disturbing.

## 1 Introduction

NLP research has repeatedly been criticized for its Anglo-centrism (Bender, 2019; Joshi et al., 2020; Held et al., 2023). Efforts to counter the hegemonic state of the field have largely focused on data collection and augmentation (e.g Team et al.; Bartelds et al., 2023), as well as cross-lingual and active learning approaches to extend large models to low-resourced contexts (e.g Ogunremi et al., 2023; Dossou et al., 2022; Ogueji et al., 2021a). However, little attention is given to the quality of large-scale datasets collected for low-resourced languages (Kreutzer et al., 2022). Further, as NLP systems are integrated into diverse social contexts, they exhibit and perpetuate harmful societal biases and may have disparate performance for different social groups (Hada et al., 2024; Savoldi et al., 2021, 2024).

However, studying societal biases is a difficult undertaking. Social structures and hierarchies are complicated dynamic systems that pull from economic, political, and cultural aspects, making it difficult to distill them down to phenomena we can measure with just a few metrics. Within NLP literature, 1) several papers that study "bias" do not define it clearly or agree on the definition (Blodgett et al., 2020), 2) benchmarks for measuring bias fall short of articulating what exactly they are measuring (Blodgett et al., 2021), and 3) most automated metrics lack actionability (Delobelle et al., 2024). All of these challenges are further complicated when we try to study bias in multilingual and multicultural settings (Talat et al., 2022).

Bias and harm due to Machine Learning (ML) systems can occur in two broad stages of an ML pipeline: data generation and model building and implementation (Suresh and Guttag, 2021). Prior work in evaluating Machine Translation models for bias has largely focused on gender bias and evaluated MT models using benchmark datasets(e.g Sewunetie et al., 2024; Wairagala et al., 2022). However, these evaluations are confined to the second stage, model building and implementation; i.e, once the model is trained, how does it perform across various social groups?

In this paper, we focus on the first stage in the ML pipeline–the data. Prior work has explored the quality of large, web-crawled multilingual datasets and found that the majority of the large-scale datasets labeled for low-resourced lan-guages may not even be in the language (Kreutzer et al., 2022). In high-resourced contexts, several works have revealed the social biases embedded in large-scale datasets (e.g Birhane et al., 2023b). We extend this literature further through a series of automated and human evaluations of MT datasets for three low-web data languages–Afan Oromo, Amharic, and Tigrinya.

---

* Authors' contributions listed in Appendix A
[1] We will release all code, data, and models upon publication

Our analysis revealed that while training data is heavily dominated by religious and political topics, benchmark datasets focus on sports, health, and news domains (Sec. 5.1). In terms of gender representation, we found that both training and benchmark datasets are heavily skewed towards the male gender with over 80% of person names being stereotypically male (Sec. 5.2), up to 72 percentage point differences between the number of male gender verbs and female gender verbs (Sec. 5.2), and stereotypical associations of occupations like engineer being consistently associated with the male gender (Sec. 5.3). Based on our results, we provide a set of recommendations for data collection and offer directions for future work evaluating datasets for low-resourced languages (Sec. 6).

## 2 Background

Here, we give background information on how the languages of study represent gender and highlight linguistic and cultural gender bias in the languages.

### 2.1 Languages of Study

**Afan Oromo:** Afan Oromo is a Cushitic language. Afan Oromo uses the Qubee alphabet, which is written with the Latin script. It has notional gender representation, where both gendered and gender-neutral expressions exist. Where gender is marked, it is indicated by the use of morphological inflections: Feminine markers include suffixes such as "-tti," "-tu," "-e," and "-itti" while masculine markers include "-ssa," "-sa," and "-a". These markers are used in nouns, adjectives, and personal names. For example, the masculine form "sooressa" refers to "rich man" while the feminine form "sooretti" refers to "rich woman."

**Amharic:** Amharic is an Afro-Semitic language. It uses the Ge'ez script for writing. Amharic is a grammatically gendered language where every noun is gendered. In addition to the female and male genders, Amharic has a gender neutral pronoun that is used as a respectful pronoun. Gender is indicated via morphological inflections: for instance, "ተማረ"(he learned) is masculine while "ተማረች"(she learned) is feminine and "ተማሩ"(they learned) can be gender neutral (respectful) or plural.

**Tigrinya:** Tigrinya is an Afro-Semitic language. Tigrinya also uses the Ge'ez script and is a grammatically gendered language. In Tigrinya, the respectful pronouns are also gendered: for instance, "መጺኡ"(he came) is masculine, "መጺኣ"(she came) is feminine. For plural and respectful terms, "መጺኦም"(they came) is used for respectful male gender, group of two or more male gender persons or a mixed group of male and female gender persons, while "መጺአን"(they came) is used for two or more female gender persons or as a respectful female gender indicator.

### 2.2 Linguistic and Cultural Gender Bias

The linguistic representation of gender may intertwine with stereotypes and biased notions. For instance, in the three languages, gender markers may be determined by the size of an object–where smaller objects are referred to with feminine gender and larger objects are referred to with masculine gender[2]. Gender bias could also be observed in the formation of proper names: proper male names may exude power and dominance while female names depict beauty, softness, and delicacy (Leyew, 2003). Additionally, gender bias in the linguistic structure of the languages may also be observed in proverbs, figurative speech, and administrative titles that perpetuate gendered stereotypes (Yadate, 2015).

Gendered stereotypes in the three languages may be observed in various online and offline texts: prior works in linguistics have demonstrated the stereotypical, more negative depictions of women in high-school textbooks of the three languages. Barkessa (2020) found that over 63% of nouns, pronouns and verbs in Afan Oromo textbook referred to the male gender and that adjectives related to the female gender included "poor", "shameful" while those related to the male gender included "resistant" and "knowledgeable." Murra (2023) found that occupational roles given to women in Amharic textbooks were "minimum and less respected." Similarly, Mesele and Asfaw (2019) analyzed Tigrinya textbooks and found "femininity is rendered invisible" in the textbooks.

---

[2]Refer to Appendix B for examples.

Despite the negative gendered stereotypes towards women reflected in the gendered representation of the languages, the languages are spoken in cultures that have some traditions to promote women. There are traditional women's institutions set in place to protect women's rights (Duressa, 2018), parts of the community that exclusively live without gender roles (Ayanaw and Alewond, 2024), festivals and holidays that exclusively celebrate women (Balehey and Balehegn, 2019), and several female historical figures that held political and leadership roles (Mamo, 2016; Bizuneh, 2001; Abebaw Ejigu, 2024). However, there is limited representation of these traditions and historical figures within written records (Abebaw Ejigu, 2024).

## 3 Related Work

Multilingual NLP research has focused on adopting language technologies to understudied and low-resourced language contexts. Particularly, MT systems are increasingly used as data augmentation and synthetic data creation methods for low-resourced languages (e.g. Singh et al., 2025; Joshi et al., 2025). However, without careful consideration of societal factors, such adoption of technology risks (1) furthering power imbalances and biases that exist in the communities and (2) importing new biases from high-resourced language contexts (Mitchell et al., 2025).

Studying bias in NLP systems requires the interrogation of language technologies as socio-technical systems (Blodgett et al., 2020). Language technologies are increasingly integrated into diverse social contexts (e.g. Vieira et al., 2021; Mehandru et al., 2022; Weissburg et al., 2025). As a result, the biases embedded in the language technologies we build can trickle down to everyday life. As Savoldi et al. (2024) demonstrated, there is a substantial difference in resources required to correct errors of MT outputs for the feminine gender as compared to the masculine gender. Further, language technologies are mostly developed under Western cultural and social contexts (Bender, 2019; Joshi et al., 2020). Hence, they may not account for the diverse social contexts of non-Western communities (Talat et al., 2022).

Sewunetie et al. (2024) evaluated gender bias in machine translation outputs between English and our three languages of focus by translating an English-centric benchmark that focused on occupational stereotypes. In this paper, we focus our evaluation on the first stage of the machine learning pipeline–data (Suresh and Guttag, 2021). Our evaluation also moves beyond occupational stereotypes and looks into person names, grammatical gender, and topics covered by the datasets.

## 4 Methods

### 4.1 Datasets

To select datasets for evaluation, we first searched for machine translation papers where the target languages are any of our three languages of focus. We collected papers from the ACL Anthology and through a general Google search. We also searched for datasets on HuggingFace (Wolf et al., 2020) using the languages and the machine translation task as filters. We found a total of 17 papers that exclusively contribute datasets in at least one of the three languages of study. In Appendix C, we provide full descriptions of all the datasets we found. Examining the domain distribution of the datasets in Fig. 3, we observe that the religious domain dataset is the most frequently used, with over 93% of the datasets containing religious text.

From the papers we found, we selected NLLB (NLLB et al., 2022), which has the largest training dataset for each language, and FLORES (Goyal et al., 2021), HornMT (Hadgu et al., 2022), and MAFAND (Adelani et al., 2022), which are evaluation benchmarks for Machine Translation. Below, we give details of the datasets we evaluated, and in Table 1, we summarize the language coverage and statistics.

**NLLB** (NLLB et al., 2022) includes parallel datasets for 200 languages, where the data is obtained from online sources through automatic data mining techniques. The data is openly available and has been used as training data by several works (e.g. Tan and Zhu, 2024; Chang et al., 2024; Xu et al., 2025).

**FLORES** (Goyal et al., 2021) is an evaluation benchmark for MT curated by collecting 3000 sentences from English Wikipedia and having the sentences human translated into 100 languages. The dataset was later extended to 200 languages, including our three languages of focus[3].

**HornMT** (Hadgu et al., 2022) is an evaluation benchmark dataset that has 2k parallel sentences for 5 languages spoken in the Horn of Africa. All three of our focus languages are included in the HornMT dataset[4].

---
[3]We accessed the dataset through HuggingFace at https://huggingface.co/datasets/facebook/flores and used the dev(n=997) and devtest(n=1012) splits combined.

| Dataset | Amharic | Tigrinya | Afan Oromo |
|---------|---------|----------|------------|
| NLLB    | 16.14M  | 1.39M    | 3.23M      |
| FLORES  | 2K      | 2K       | 2K         |
| HornMT  | 2K      | 2K       | 2K         |
| MAFAND  | 1.94K   | -        | -          |

Table 1: Datasets evaluated along with the number of parallel sentences and language coverage.

**MAFAND** (Adelani et al., 2022) is an evaluation benchmark dataset for 21 African languages. The dataset is curated from the news domain. Of our three target languages, only Amharic is included in MAFAND, with 1.94k parallel sentences with English.

### 4.2 Evaluation Methods

Below, we detail our supervised and unsupervised methods, as well as human evaluation schemes[5].

#### 4.2.1 Topic Modeling

Topic modeling refers to the task of identifying themes across documents. For our purposes, we used topic modeling to uncover the themes in each of the datasets and to understand if there are any correlations between the topics and the gender representation in the data. We used Latent Dirichlet Allocation (LDA) (Blei et al., 2003), which is a popular method to date (Sobchuk and Šeļa, 2024; Lucy et al., 2025).

**Model Training** We trained LDA models on unigrams for each language and each dataset using MALLET (McCallum, 2002) via the *Little Mallet Wrapper* API[6]. For the benchmark datasets, we used 5 topics and for the NLLB dataset, we trained with 50 topics for each language. Furthermore, for the NLLB dataset, we trained a total of three topic models per language:

- General topic model: We trained an LDA model on the full dataset preprocessed as described above.

- Gender-Specific topic model: We prepared a set of gender-specific keywords and filtered sentences from the full dataset into feminine and masculine sentences. We then trained LDA models for each group of sentences. Further details are in Appendix D

**Evaluation** Once we had the topics, native speakers of each language manually inspected the salient terms in each topic and provided a summary of each topic.

#### 4.2.2 Morphological Analysis

As described in Sec. 2.1, gender in verbs for the three languages is indicated through the addition of suffixes and prefixes. Hence, we used morphological analysis to understand the gendered representation of verbs in the MT datasets.

**Morphological Analyzer** We used HornMorpho (Gasser, 2011), which is a rule-based morphological analyzer built using a finite-state transducer (FTS). HornMorpho relies on explicit rules and a finite lexicon, which is compiled from online dictionaries for each language. HornMorpho includes all three of our languages of focus. When analyzing a verb, HornMorpho outputs subject agreement, including the gender of the subject of the verb. Further details can be found in Appendix F.

**Evaluation** We automatically counted the gender identified by HornMorpho for each of the verbs identified. We also manually verified and analyzed the top 20 most frequent verbs for each dataset for each language.

#### 4.2.3 Named Entity Recognition

Named Entity Recognition (NER) is a supervised NLP task where models are trained to extract named entities–such as person names, dates, and locations–from a given text. For our study, we used NER models to extract person names from the datasets we are evaluating. We then performed an analysis to understand the gender distribution of people included in the datasets.

**Model Training** For Tigrinya and Afan Oromo, we finetuned AfriBERTa-base and AfriBERTa-large, respectively, as both languages are included in the pre-training of the AfriBERTa model. For Amharic, we chose the AfroXLMR-large (Alabi et al., 2022a) model, which has a better reported performance for NER compared to the AfriBERTa model for Amharic (Alabi et al., 2022a). Since AfroXLMR does not include Tigrinya and Afan Oromo in pre-training, we refrained from using it for the two languages. Further finetuning details and datasets used to finetune the models are available in Appendix E.

---

[4]https://github.com/asmelashteka/HornMT
[5]For further details on methods (e.g, pre-processing strategies), please refer to the Appendix D- G.
[6]https://github.com/maria-antoniak/little-mallet-wrapper

**Evaluation** Since the number of names identified from each dataset was large, we used a stratified sampling approach to select a representative set for analysis. We split the list of names into three strata, in the first one, we included all names that appeared in the dataset at least 100 times. Then, from the remaining list of names, we randomly selected 100 names that had a frequency range [101,20], and finally randomly sampled 100 names that had a frequency range [19, 1]. Native speakers of each language then looked through the samples and labeled the names as stereotypically feminine, masculine, or gender neutral. We also labeled for the category of the names as religious, political figure, and regular name.

### 4.2.4 Masked Language Modeling

We used Masked Language Models (MLMs) to understand the contextual representation of biases encoded in the datasets.

**Model Training** We used AfriBERTa-small model (Ogueji et al., 2021b) and mBERT (Devlin et al., 2018) in our experiments. AfriBERTa contains all three languages, while mBERT does not include any of the three languages in pretraining. We used the NLLB dataset to fine-tune AfriBERTa and mBERT models for each of the three languages, resulting in a total of 6 fine-tuned MLM models[7].

**Evaluation** We prepared an evaluation dataset with 22 seed sentences using stereotypical occupations and adjectives. We then prepared clozes by masking the nouns, adjectives, and verbs in each sentence, which resulted in 376 masked sentences. We marked the expected gender of the masked token and used the MLMs to predict the masked token. We then compared the gender of the predicted masked token to the expected gender of the masked token. We also got the top 5 predictions from the MLMs and counted the gender in the predicted tokens. We prepared this evaluation dataset for each of the three target languages. We provide more details and examples in Appendix G. Since AfriBERTa already includes the three languages in pre-training, we performed our evaluation before and after finetuning for AfriBERTa and report on the changes we observed.

## 5 Results

In this section, we present our findings from the evaluations we described in the previous sections.

### 5.1 Identified Topics

As described in Sec. 4.2.1, we trained topic models for each dataset in each language. We then looked at the salient terms for each topic and assigned a label for what the overall category represents.

**Political and religious topics dominate the NLLB dataset for all three languages.** We observe that for all three languages, of the 50 topics identified for the NLLB dataset, religious topics accounted for the majority (8%, 12%, 20% for Afan Oromo, Amharic, and Tigrinya, respectively) while political topics accounted for 22%, 8%, 14% of the topics for Afan Oromo, Amharic, and Tigrinya, respectively. Other topics accounted for less than 6% each, with some having only a single occurrence per language. Specific to each language, we observed that the Afan Oromo NLLB dataset had a few topics on language and cultural representation. We also observed that the datasets for one of the languages had heavy representation of topics related to refugees and immigration[8].

**Gendered terms appear in family roles, legal matters, and healthcare topics for the NLLB dataset.** We observed male and female gendered terms in the topic for family role, including terms like "mother," "wife," "father," and "man." For healthcare topic in Amharic, we observe that "women" is among the salient terms, along with terms like "vaccine." Topics on finance and politics are mostly gender neutral, and focus more on domain specific terms like "vote," and "exchange rate." For Afan Oromo, we observed that the term "abbaa"(father) appears along with other salient terms for a topic on legal matters. For Tigrinya, we identified a topic where the majority of the salient terms were dedicated to women and gender issues. Table 4 gives examples in detail.

**Topics in sentences with female and male keywords include reproductive health, sports activities, and family roles.** Looking at the salient

---

[7] While we tried finetuning MLMs for the benchmark datasets, the size of the datasets was too small to effectively train the model. Hence, we only performed full evaluation and training on the MLMs finetuned using the NLLB dataset.

[8] Following guidance from Kirk et al. (2022), we refrain from naming the language to avoid negative connotations to the language or language speakers.

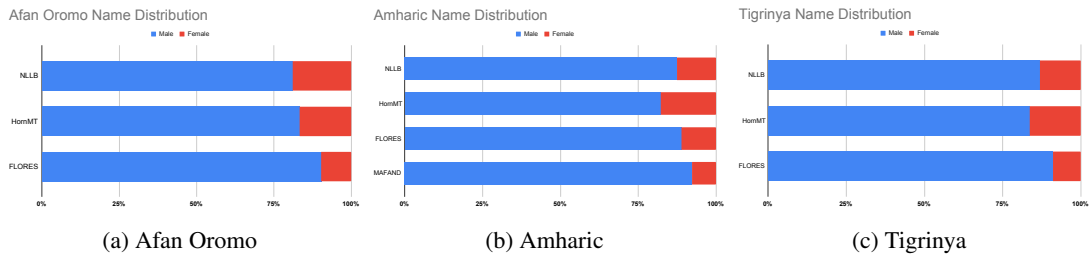

Figure 1: Percentage of stereotypically female and stereotypically male person names in Afan Oromo, Amharic, and Tigrinya datasets.

terms of the topics trained on sentences with female and male keywords from the NLLB dataset, we found that for the sentences with female keywords, topics include reproduction, with salient terms like "reproductive organ" and "body part." There are also topics with salient terms like "wife," "husband," "marriage," and "elders" which refer to family roles and social structures. Conversely, for the sentences with male keywords, we found that gendered terms appeared for topics on social roles and family roles. In the topic with social roles for Amharic, one of the top words was a derogatory term used to refer to female sex workers. Table 5 gives examples of topics for female gender sentences and Table 6 gives examples of topics for male gender sentences.

**Salient terms include code-mixed words from other languages.** We observed a few instances where the salient terms identified were not in the language of focus. This was most notable for Tigrinya and Afan Oromo, with the latter having more code-mixed topics. We provide examples in Table 4. For Afan Oromo, we further investigated the dataset by manually inspecting ~300 sentences. As Fig. 4 shows, over 40% of the manually inspected sentences were not in the language.

**Benchmark datasets mostly have news, health, and sport-related topics, with some that portray negative stereotypes against women.** Since HornMT and FLORES datasets are parallel across the three languages, we found that the topics identified cover similar grounds for all three languages. We found that most of the topics in benchmark datasets are centered on news. HornMT dataset had one topic on healthcare, which included salient terms like "virus" and "epidemic." On the other hand, MAFAND dataset, which only has Amharic, included topics with negative portrayals of women with salient terms like "hot girls," "university campus," and "for rich people" occurring in a single topic. The same topic also included salient terms like "monster" and the name of one ethnic group. Table 7 gives examples of topics and salient terms for the benchmark datasets.

### 5.2 Subjects and Verbs in MT datasets

In this section, we provide the results of our NER and Morphological Analysis experiments described above.

**Stereotypically female names account for less than 20% of the total names in MT datasets** As described in Sec. 4.2.3, we labeled the person names identified by the NER models we trained for each language. We find that the percentage of female names could be as low as 7.74% for MAFAND in Amharic, and up to 18.98% for NLLB in Afan Oromo. Across the benchmark datasets, we see that HornMT has a higher percentage of female names as compared to FLORES and MAFAND. We present the visualization of these results in Fig. 1.

**Most frequent names in NLLB are primarily names from religious texts, while those of benchmarks are regular names.** For the names that appeared at least 100 times, we labeled them into three categories: religious, political figure, and regular name. We find that most of the names from the NLLB dataset are religious names for Amharic and Tigrinya. From the benchmark datasets, we observe that HornMT has more political figure names compared to FLORES and MAFAND. We also observe that in all three benchmark datasets, regular names account for above 50% of the names in the top 100 most frequent names for all three languages. Further, the benchmark datasets have a small percentage of religious names in all three languages. Fig. 5 presents an illustration of the percentage distribution.

**Most verbs in the datasets have male grammatical gender.** Fig. 2 shows the distribution of verbs with male gender inflections vs those with female gender inflections. We observed that the majority of the verbs have male gender inflection for all datasets in all languages, with the gaps being more pronounced for Tigrinya and Amharic, which are grammatically gendered (see Sec. 2.1).

### 5.3 Contextual Bias

Here, we describe our results for the evaluation of the MLMs finetuned with the NLLB dataset as described in Sec. 4.2.4.

**Wrong predictions are more likely to occur for masked tokens with female expected gender for all languages, in most models.** In Table 14, we note that for all three languages and most models, wrong tokens were predicted for data points where the expected gender was female as compared to the data points with male expected gender. For instance, 2.53% of the predicted tokens were wrong for the Male expected gender for Tigrinya with the AfriBERTa model finetuned with NLLB dataset, compared to 6.49% of wrong predictions for the female expected gender with the same model.

**Finetuning with NLLB data decreased wrong predictions and increased male gender predictions for Afan Oromo and Amharic, regardless of the expected gender.** As described in Sec. 4.2.4, we experimented with AfriBERTa-Plain, where we run the model as is on our evaluation dataset, and AfriBERTa-Finetuned, where we finetuned the AfriBERTa model with the NLLB dataset for each language. Since AfriBERTa includes all three languages in pre-training, the AfriBERTa-Finetuned results tell us how the NLLB dataset shifts the distribution of the pre-trained model. As can be seen in Table 14, finetuning with NLLB consistently decreased the wrong predictions for all languages, regardless of the expected gender. For instance, 19.74% of the predicted tokens were incorrect in AfriBERTa-Plain for Amharic; this was reduced to 10.53% with AfriBERTa-Finetuned. Further, the AfriBERTa-Finetuned model outputs increased the predicted tokens with male gender connotations for Afan Oromo and Amharic; the percentage of predicted tokens with male gender connotations increased by 12, 17.6 percentage points for Amharic, Afan Oromo respectively. However, the increase in predicted tokens with male gender connotations happened even when the expected gender is female. On the other hand, finetuning eliminated cases where the predicted token has female gender connotation while the expected gender is male. Tigrinya was an exception to this, where finetuning with NLLB decreased the male gender predicted token for female gender expected token by a 33.85 percentage point.

**Amharic had the highest percentage of wrong predictions, and Afan Oromo had the highest rate of neutral predictions.** While Amharic had the largest dataset from our three languages of study, the predictions of the MLMs had the highest percentage of wrong predictions compared to Afan Oromo and Tigrinya (see Table 14). Further, when the expected gender is female, all MLM model predictions had a higher percentage of predicted tokens with male connotations for Amhairc as compared to the other languages. For example, for mBERT-finetuned, 46.05% of the model predictions had male connotations for Amharic. However, it is important to note that Afan Oromo had the highest percentage of neutral predictions, regardless of the expected gender. This could be due to the grammatical nature of the language as described in Sec. 2

**Some adjectives and occupations had a strong correlation with a particular gender.** In both Afan Oromo and Amharic, we found that the adjectives for "beautiful" and "emotional/nagging" are strongly correlated with the female gender. When those adjectives are used, the model predictions frequently output tokens with female gender connotations. When the adjectives were used in a male gender context, the models' predictions output tokens that did not make sense in the context of the sentence. Similarly, certain occupations were consistently associated with the male gender: "business owner," "lawyer," and "cleaner" in Afan Oromo and "engineer" in Tigrinya were consistently associated with the male gender, while "nurse" was consistently associated with the female gender for Amharic.

## 6 Discussion

**More Data, More Problems?** Our work reveals that the topics for training data and benchmark data do not always align (Sec. 5.1). Further, we found that languages with less data contain code-mixing and sentences in other languages (Sec. 5.1). More research into Language ID tools (e.g. Gaim et al., 2022) as well as human-centric and participatory approaches (e.g. Nekoto et al., 2020) may offer better alternatives for future data collection schemes. While collecting more data may reduce the quality issues related to language mixing, it may also introduce its own set of problems. We found that the language with the most data contained more toxic and harmful topics (Sec. 5.1). This finding aligns with prior work that evaluated textual descriptions of computer vision datasets and found that hateful content increased with the dataset's scale (Birhane et al., 2023a). Hence, without careful evaluation of our datasets and data collection practices, more data may lead to models that coherently output harmful content.

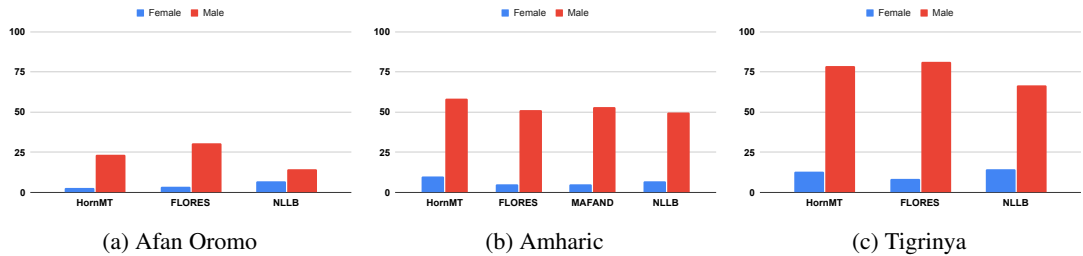

Figure 2: Percentage distribution of grammatical gender of verbs in Afan Oromo, Amharic and Tigrinya datasets.

**The Persistence of Gender Bias in NLP Systems**
Gender bias has been studied extensively in NLP and particularly in MT literature (e.g Hada et al., 2024; Zhao et al., 2018; Wairagala et al., 2022; Savoldi et al., 2021). In fact, prior work has called for expanding our evaluation of bias beyond gender (Talat et al., 2022). However, as our findings revealed, the problem persists. We found a stark skew in the data to the male gender in terms of names of persons and grammatical gender of verbs (Sec. 5.2). Our analysis also revealed a stereotypical correlation between certain adjectives and occupations that are consistently associated with one gender over the other (Sec. 5.3). Despite the findings from linguistic studies demonstrating the linguistic and cultural embedding of gender bias (see Sec. 2.2), we observed the same phenomenon in MT datasets. We argue that this is largely due to how bias evaluation is treated as an after-the-fact undertaking. Instead, we call for data collection schemes that incorporate bias investigation by design. As we have discussed in Sec. 2.2, there are several cultures and stories about women within the communities; it is a matter of actively representing them in our datasets.

**What Do MT Benchmarks Measure?** Our analysis revealed that benchmark datasets for MT mostly covered news, sport and health-related topics (sec. 5.1). While the benchmark datasets evaluated in this paper are not exclusively designed to measure gender bias, they should not perpetuate it. We find that the text included in the benchmarks is skewed towards the male gender and, at times, includes harmful content. We again call for the integration of bias evaluation in the design stage of curating benchmarks.

**Implications for Models and Downstream Applications.** As we have discussed in Sec. 1, we focused our analysis on the first stage in the machine learning pipeline. Prior work evaluated two machine translation models for gender bias, including the NLLB model, which is trained with the NLLB dataset we evaluated (Sewunetie et al., 2024). Unsurprisingly, Sewunetie et al. (2024) found that the NLLB model showed high levels of bias against the female gender, based on a translated WinoBias (Zhao et al., 2018) benchmark. Our analysis supplements this finding by giving insights into the biases embedded in the training data. Our findings also indicated an over-representation of certain topics in the dataset for some languages, which could have downstream effects. We focused our analysis on MT datasets as MT models are not only used as standalone models, but also for generating data for other models and applications (e.g. Singh et al., 2025; Joshi et al., 2025). However, our approach and analysis can be extended to any other dataset and NLP task.

## 7 Conclusion

Through a series of human and automated evaluations, our work sheds light on the state of machine translation datasets for three low-web data languages. Our findings demonstrated mismatches in the domain of training and benchmark datasets, and highlighted a heavy skew towards the male gender. We hope our findings inspire more cautious curation and evaluation of datasets for low-resourced languages.

**Limitations**

Our work has several limitations. First, although we are evaluating MT datasets, we focused on the data in the low-resourced languages; we did not evaluate the parallel English sentences. However, we were interested in understanding the linguistic, cultural, and grammatical nature of gender bias in the three languages within MT datasets and the quality of the target language side of the data. Future work could, for instance, incorporate an evaluation of the parallel sentence and check if they are correctly aligned. Further, as we have discussed throughout the paper, we focused on evaluating the data; bias could also be introduced during the model training and deployment phase (Suresh and Guttag, 2021). Future work can extend our study with an evaluation of MT models that specifically address the pitfalls we identified: i.e, person names, adjectives, and gendered verbs. Further, we had to apply sampling methods in our NER, Morphological Analysis, and qualitative evaluations due to the size of the dataset, particularly for NLLB. We describe our methods for sampling in Sec. 4 and Appendix D- G.

**Ethical Considerations**

Although they are publicly available, the datasets we evaluated include personally identifiable information such as names of individuals. We refrained from including personal names from the datasets in our paper. Further, we found some correlations between certain topics and individual languages, which may perpetuate negative stereotypes of speakers of the languages (Sec. 5.1). We followed guidance from Kirk et al. (2022) and omitted naming the languages where such risks might manifest. Similarly, we omitted naming an ethnic group that was a salient term in a topic with toxic terms. Additionally, the datasets include large amounts of religious data, which requires careful ethical consideration when used in NLP (Hutchinson, 2024). For instance, we refrained from giving examples of religious salient terms in our topic modeling tables (Table 4- 6) so as not to present text from religious books along with toxic and harmful terms. Additionally, we labeled the names of individuals for gender and category. It should be noted that we relied on stereotypical gender associations to perform the labeling.

**Acknowledgments**

We thank ACM FAccT for funding this project. We also thank the various paid translators who helped us in this project: Galatoomaa, እናመሰግናለን, የቐኒ-የለና። ንኣይተ መንገሻ ያዮ ኣብ ጇንጃ ትግርኛ ምትርጓም ኣብ ዝገበርካልና ሓገዝ ካብ ልቢ ነመስግን። (We thank Mr. Mengesha Yayo for his input on the Tigrinya analysis and translations). We also thank Michael Gasser for providing insights and support on HornMorpho.

**References**

Abdo Ababor. 2021. Boosting afaan oromo named entity recognition with multiple methods. *International Journal of Information Engineering & Electronic Business*, 13(5).

Solomon Teferra Abate, Michael Melese, Martha Yifiru Tachbelie, Million Meshesha, Solomon Atinafu, Wondwossen Mulugeta, Yaregal Assabie, Hafte Abera, Binyam Ephrem, Tewodros Abebe, Wondimagegnhue Tsegaye, Amanuel Lemma, Tsegaye Andargie, and Seifedin Shifaw. 2018. Parallel corpora for bi-lingual English-Ethiopian languages statistical machine translation. In *Proceedings of the 27th International Conference on Computational Linguistics*, pages 3102–3111, Santa Fe, New Mexico, USA. Association for Computational Linguistics.

Tesfaw Abebaw Ejigu. 2024. The overlooked roles of women in the patriotic resistance movement in Bure Damot, 19361941. *Cogent Arts & Humanities*, 11(1):2390786. Publisher: Cogent OA _eprint: https://doi.org/10.1080/23311983.2024.2390786.

D. Adelani, Jade Abbott, Graham Neubig, Daniel D'Souza, Julia Kreutzer, Constantine Lignos, Chester Palen-Michel, Happy Buzaaba, Shruti Rijhwani, Sebastian Ruder, Stephen Mayhew, Israel Abebe Azime, S. Muhammad, Chris C. Emezue, Joyce Nakatumba-Nabende, Perez Ogayo, Anuoluwapo Aremu, Catherine Gitau, Derguene Mbaye, J. Alabi, Seid Muhie Yimam, Tajuddeen R. Gwadabe, Ignatius Ezeani, Rubungo Andre Niyongabo, Jonathan Mukiibi, V. Otiende, Iroro Orife, Davis David, Samba Ngom, Tosin P. Adewumi, Paul Rayson, Mofetoluwa Adeyemi, Gerald Muriuki, Emmanuel Anebi, C. Chukwuneke, N. Odu, Eric Peter Wairagala, S. Oyerinde, Clemencia Siro, Tobius Saul Bateesa, Temilola Oloyede, Yvonne Wambui, Victor Akinode, Deborah Nabagereka, Maurice Katusiime, Ayodele Awokoya, Mouhamadane Mboup, D. Gebreyohannes, Henok Tilaye, Kelechi Nwaike, Degaga Wolde, Abdoulaye Faye, Blessing Sibanda, Orevaoghene Ahia, Bonaventure F. P. Dossou, Kelechi Ogueji, Thierno Ibrahima Diop, A. Diallo, Adewale

Akinfaderin, T. Marengereke, and Salomey Osei. 2021a. Masakhaner: Named entity recognition for african languages. *ArXiv*, abs/2103.11811.

David Adelani, Jesujoba Alabi, Angela Fan, Julia Kreutzer, Xiaoyu Shen, Machel Reid, Dana Ruiter, Dietrich Klakow, Peter Nabende, Ernie Chang, Tajuddeen Gwadabe, Freshia Sackey, Bonaventure F. P. Dossou, Chris Emezue, Colin Leong, Michael Beukman, Shamsuddeen Muhammad, Guyo Jarso, Oreen Yousuf, Andre Niyongabo Rubungo, Gilles Hacheme, Eric Peter Wairagala, Muhammad Umair Nasir, Benjamin Ajibade, Tunde Ajayi, Yvonne Gitau, Jade Abbott, Mohamed Ahmed, Millicent Ochieng, Anuoluwapo Aremu, Perez Ogayo, Jonathan Mukiibi, Fatoumata Ouoba Kabore, Godson Kalipe, Derguene Mbaye, Allahsera Auguste Tapo, Victoire Memdjokam Koagne, Edwin Munkoh-Buabeng, Valencia Wagner, Idris Abdulmumin, Ayodele Awokoya, Happy Buzaaba, Blessing Sibanda, Andiswa Bukula, and Sam Manthalu. 2022. A Few Thousand Translations Go a Long Way! Leveraging Pre-trained Models for African News Translation. In *Proceedings of the 2022 Conference of the North American Chapter of the Association for Computational Linguistics: Human Language Technologies*, pages 3053–3070, Seattle, United States. Association for Computational Linguistics.

David Ifeoluwa Adelani, Jade Abbott, Graham Neubig, Daniel Dsouza, Julia Kreutzer, Constantine Lignos, Chester Palen-Michel, Happy Buzaaba, Shruti Rijhwani, Sebastian Ruder, Stephen Mayhew, Israel Abebe Azime, Shamsuddeen H. Muhammad, Chris Chinenye Emezue, Joyce Nakatumba-Nabende, Perez Ogayo, Aremu Anuoluwapo, Catherine Gitau, Derguene Mbaye, Jesujoba Alabi, Seid Muhie Yimam, Tajuddeen Rabiu Gwadabe, Ignatius Ezeani, Rubungo Andre Niyongabo, Jonathan Mukiibi, Verrah Otiende, Iroro Orife, Davis David, Samba Ngom, Tosin Adewumi, Paul Rayson, Mofetoluwa Adeyemi, Gerald Muriuki, Emmanuel Anebi, Chiamaka Chukwuneke, Nkiruka Odu, Eric Peter Wairagala, Samuel Oyerinde, Clemencia Siro, Tobius Saul Bateesa, Temilola Oloyede, Yvonne Wambui, Victor Akinode, Deborah Nabagereka, Maurice Katusiime, Ayodele Awokoya, Mouhamadane MBOUP, Dibora Gebreyohannes, Henok Tilaye, Kelechi Nwaike, Degaga Wolde, Abdoulaye Faye, Blessing Sibanda, Orevaoghene Ahia, Bonaventure F. P. Dossou, Kelechi Ogueji, Thierno Ibrahima DIOP, Abdoulaye Diallo, Adewale Akinfaderin, Tendai Marengereke, and Salomey Osei. 2021b. MasakhaNER: Named Entity Recognition for African Languages. *Transactions of the Association for Computational Linguistics*, 9:1116–1131.

Sisay Adugna and Andreas Eisele. 2010. English — Oromo machine translation: An experiment using a statistical approach. In *Proceedings of the Seventh International Conference on Language Resources and Evaluation (LREC'10)*, Valletta, Malta. European Language Resources Association (ELRA).

Jesujoba Alabi, David Ifeoluwa Adelani, Marius Mosbach, and Dietrich Klakow. 2022a. Adapting Pre-trained Language Models to African Languages via Multilingual Adaptive Fine-Tuning. *ACL Anthology*, pages 4336–4349.

Jesujoba O. Alabi, David Ifeoluwa Adelani, Marius Mosbach, and Dietrich Klakow. 2022b. Adapting pre-trained language models to African languages via multilingual adaptive fine-tuning. In *Proceedings of the 29th International Conference on Computational Linguistics*, pages 4336–4349, Gyeongju, Republic of Korea. International Committee on Computational Linguistics.

Jisun An, Haewoon Kwak, and Yong-Yeol Ahn. 2018. SemAxis: A Lightweight Framework to Characterize Domain-Specific Word Semantics Beyond Sentiment. In *Proceedings of the 56th Annual Meeting of the Association for Computational Linguistics (Volume 1: Long Papers)*, pages 2450–2461, Melbourne, Australia. Association for Computational Linguistics.

Yeabsira Asefa Ashengo, Rosa Tsegaye Aga, and Surafel Lemma Abebe. 2021. Context based machine translation with recurrent neural network for englishamharic translation. *Machine Translation*, 35:19 – 36.

Gashaw Ayanaw and Abebe Alewond. 2024. Mainstreaming gender equality: A comparative analysis on gender roles of the Awura Amba and the nearby community, Amhara Regional State, Ethiopia. *Social Sciences & Humanities Open*, 10:101149.

Selam Balehey and Mulubrhan Balehegn. 2019. The Art, Aesthetics and Gender Significance of Ashenda girls' Festival in Tigray, Northern Ethiopia.

Adugna Barkessa. 2020. A Critical Discourse Analysis of Gendered Representations in Afaan Oromoo Textbook. *East African Journal of Social Sciences and Humanities*, 5(1):19–36.

Martijn Bartelds, Nay San, Bradley McDonnell, Dan Jurafsky, and Martijn Wieling. 2023. Making More of Little Data: Improving Low-Resource Automatic Speech Recognition Using Data Augmentation. In *Proceedings of the 61st Annual Meeting of the Association for Computational Linguistics (Volume 1: Long Papers)*, pages 715–729, Toronto, Canada. Association for Computational Linguistics.

Tadesse Destaw Belay, Atnafu Lambebo Tonja, Olga Kolesnikova, Seid Muhie Yimam, Abinew Ali Ayele, Silesh Bogale Haile, Grigori Sidorov, and Alexander Gelbukh. 2022. The effect of normalization for bi-directional amharic-english neural machine translation. *Preprint*, arXiv:2210.15224.

Emily Bender. 2019. The #BenderRule: On Naming the Languages We Study and Why It Matters.

Zemicheal Berihu, Gebremariam Mesfin, Mulugeta Atsibaha, and Tor-Morten Grønli. 2020. Enhancing bi-directional english-tigrigna machine translation using hybrid approach.

Yohanens Biadgligne and Kamel Smaïli. 2021. Parallel corpora preparation for english-amharic machine translation. In *Advances in Computational Intelligence*, pages 443–455, Cham. Springer International Publishing.

Yohannes Biadgligne and Kamel Smaili. 2022. Offline corpus augmentation for english-amharic machine translation. In *2022 5th International Conference on Information and Computer Technologies (ICICT)*, pages 128–135.

Abeba Birhane, Vinay Prabhu, Sang Han, and Vishnu Naresh Boddeti. 2023a. On Hate Scaling Laws For Data-Swamps. *arXiv preprint*. ArXiv:2306.13141 [cs].

Abeba Birhane, Vinay Prabhu, Sang Han, Vishnu Naresh Boddeti, and Alexandra Sasha Luccioni. 2023b. Into the LAIONs Den: Investigating Hate in Multimodal Datasets.

Belete Bizuneh. 2001. Women in Ethiopian History: A Bibliographic Review. *Northeast African Studies*, 8(3):7–32. Publisher: Michigan State University Press.

David M. Blei, Andrew Y. Ng, and Michael I. Jordan. 2003. Latent dirichlet allocation. *J. Mach. Learn. Res.*, 3(null):993–1022.

Su Lin Blodgett, Solon Barocas, Hal Daumé Iii, and Hanna Wallach. 2020. Language (Technology) is Power: A Critical Survey of Bias in NLP. In *Proceedings of the 58th Annual Meeting of the Association for Computational Linguistics*, pages 5454–5476, Online. Association for Computational Linguistics.

Su Lin Blodgett, Gilsinia Lopez, Alexandra Olteanu, Robert Sim, and Hanna Wallach. 2021. Stereotyping Norwegian Salmon: An Inventory of Pitfalls in Fairness Benchmark Datasets. In *Proceedings of the 59th Annual Meeting of the Association for Computational Linguistics and the 11th International Joint Conference on Natural Language Processing (Volume 1: Long Papers)*, pages 1004–1015, Online. Association for Computational Linguistics.

Sisay Adugna Chala, Bekele Debisa, Amante Diriba, Silas Getachew, Chala Getu, and Solomon Shiferaw. 2021. Crowdsourcing parallel corpus for english-oromo neural machine translation using community engagement platform. *ArXiv*, abs/2102.07539.

Tyler A. Chang, Catherine Arnett, Zhuowen Tu, and Benjamin K. Bergen. 2024. Goldfish: Monolingual Language Models for 350 Languages. *arXiv preprint*. ArXiv:2408.10441 [cs].

Pieter Delobelle, Giuseppe Attanasio, Debora Nozza, Su Lin Blodgett, and Zeerak Talat. 2024. Metrics for What, Metrics for Whom: Assessing Actionability of Bias Evaluation Metrics in NLP. In *Proceedings of the 2024 Conference on Empirical Methods in Natural Language Processing*, pages 21669–21691, Miami, Florida, USA. Association for Computational Linguistics.

Jacob Devlin, Ming-Wei Chang, Kenton Lee, and Kristina Toutanova. 2018. BERT: Pre-training of Deep Bidirectional Transformers for Language Understanding. *arXiv*.

Bonaventure F. P. Dossou, Atnafu Lambebo Tonja, Oreen Yousuf, Salomey Osei, Abigail Oppong, Iyanuoluwa Shode, Oluwabusayo Olufunke Awoyomi, and Chris Emezue. 2022. AfroLM: A Self-Active Learning-based Multilingual Pretrained Language Model for 23 African Languages. In *Proceedings of the Third Workshop on Simple and Efficient Natural Language Processing (SustaiNLP)*, pages 52–64, Abu Dhabi, United Arab Emirates (Hybrid). Association for Computational Linguistics.

Endalkachew Lelisa Duressa. 2018. Review on the Historical Prospective of Siiqqee: The Ancient African Indigenous Institution for Empowering Women. *Journal of Philosophy, Culture and Religion*, 36.

Chris Emezue, Hellina Nigatu, Cynthia Thinwa, Helper Zhou, Shamsuddeen Muhammad, Lerato Louis, Idris Abdulmumin, Samuel Oyerinde, Benjamin Ajibade, Olanrewaju Samuel, Oviawe Joshua, Emeka Onwuegbuzia, Handel Emezue, Ifeoluwatayo A. Ige, Atnafu Lambebo Tonja, Chiamaka Chukwuneke, Bonaventure F. P. Dossou, Naome A. Etori, Mbonu Chinedu Emmanuel, Oreen Yousuf, Kaosarat Aina, and Davis David. 2022. The African Stopwords project: curating stopwords for African languages. *AfricaNLP*.

Fitsum Gaim, Wonsuk Yang, and Jong C. Park. 2022. GeezSwitch: Language Identification in Typologically Related Low-resourced East African Languages. In *Proceedings of the Thirteenth Language Resources and Evaluation Conference*, pages 6578–6584, Marseille, France. European Language Resources Association.

Michael Gasser. 2011. HornMorpho: a system for morphological processing of Amharic, Oromo, and Tigrinya. *ResearchGate*.

Andargachew Mekonnen Gezmu, Andreas Nürnberger, and Tesfaye Bayu Bati. 2022. Extended parallel corpus for Amharic-English machine translation. In *Proceedings of the Thirteenth Language Resources and Evaluation Conference*, pages 6644–6653, Marseille, France. European Language Resources Association.

Naman Goyal, Cynthia Gao, Vishrav Chaudhary, Peng-Jen Chen, Guillaume Wenzek, Da Ju, Sanjana Krishnan, Marc'Aurelio Ranzato, Francisco Guzman, and Angela Fan. 2021. The FLORES-101 Evaluation Benchmark for Low-Resource and Multilingual Machine Translation. *arXiv preprint*. ArXiv:2106.03193 [cs].

Rishav Hada, Safiya Husain, Varun Gumma, Harshita Diddee, Aditya Yadavalli, Agrima Seth, Nidhi Kulkarni, Ujwal Gadiraju, Aditya Vashistha, Vivek Seshadri, and Kalika Bali. 2024. Akal Badi ya Bias: An Exploratory Study of Gender Bias in Hindi Language Technology. In *Proceedings of the 2024 ACM*

*Conference on Fairness, Accountability, and Transparency*, FAccT '24, pages 1926–1939, New York, NY, USA. Association for Computing Machinery.

Asmelash Teka Hadgu, Gebrekirstos G. Gebremeskel, and Abel Aregawi. 2022. Machine translation benchmark dataset for languages in the horn of africa.

William Held, Camille Harris, Michael Best, and Diyi Yang. 2023. A Material Lens on Coloniality in NLP. *arXiv preprint*. ArXiv:2311.08391 [cs].

Ben Hutchinson. 2024. Modeling the Sacred: Considerations when Using Religious Texts in Natural Language Processing. In *Findings of the Association for Computational Linguistics: NAACL 2024*, pages 1029–1043, Mexico City, Mexico. Association for Computational Linguistics.

Pratik Joshi, Sebastin Santy, Amar Budhiraja, Kalika Bali, and Monojit Choudhury. 2020. The State and Fate of Linguistic Diversity and Inclusion in the NLP World. In *Proceedings of the 58th Annual Meeting of the Association for Computational Linguistics*, pages 6282–6293, Online. Association for Computational Linguistics.

Raviraj Joshi, Kanishk Singla, Anusha Kamath, Raunak Kalani, Rakesh Paul, Utkarsh Vaidya, Sanjay Singh Chauhan, Niranjan Wartikar, and Eileen Long. 2025. Adapting Multilingual LLMs to Low-Resource Languages using Continued Pre-training and Synthetic Corpus. *arXiv preprint*. ArXiv:2410.14815 [cs].

Hannah Kirk, Abeba Birhane, Bertie Vidgen, and Leon Derczynski. 2022. Handling and Presenting Harmful Text in NLP Research. In *Findings of the Association for Computational Linguistics: EMNLP 2022*, pages 497–510, Abu Dhabi, United Arab Emirates. Association for Computational Linguistics.

Tsegay Kiros. 2020. Statistical machine translator for english to tigrigna translation.

Julia Kreutzer, Isaac Caswell, Lisa Wang, Ahsan Wahab, Daan van Esch, Nasanbayar Ulzii-Orshikh, Allahsera Tapo, Nishant Subramani, Artem Sokolov, Claytone Sikasote, Monang Setyawan, Supheakmungkol Sarin, Sokhar Samb, Benoît Sagot, Clara Rivera, Annette Rios, Isabel Papadimitriou, Salomey Osei, Pedro Ortiz Suarez, Iroro Orife, Kelechi Ogueji, Andre Niyongabo Rubungo, Toan Q. Nguyen, Mathias Müller, André Müller, Shamsuddeen Hassan Muhammad, Nanda Muhammad, Ayanda Mnyakeni, Jamshidbek Mirzakhalov, Tapiwanashe Matangira, Colin Leong, Nze Lawson, Sneha Kudugunta, Yacine Jernite, Mathias Jenny, Orhan Firat, Bonaventure F. P. Dossou, Sakhile Dlamini, Nisansa de Silva, Sakine Çabuk Ball, Stella Biderman, Alessia Battisti, Ahmed Baruwa, Ankur Bapna, Pallavi Baljekar, Israel Abebe Azime, Ayodele Awokoya, Duygu Ataman, Orevaoghene Ahia, Oghenefego Ahia, Sweta Agrawal, and Mofetoluwa Adeyemi. 2022. Quality at a Glance: An Audit of Web-Crawled Multilingual Datasets. *Transactions of the Association for Computational Linguistics*, 10:50–72.

Zelalem Leyew. 2003. Amharic Personal Nomenclature: A Grammar and Sociolinguistic Insight.

Li Lucy, Camilla Griffiths, Sarah Levine, Jennifer L. Eberhardt, Dorottya Demszky, and David Bamman. 2025. Tell, Don't Show: Leveraging Language Models' Abstractive Retellings to Model Literary Themes. *ACL Anthology*, pages 22585–22610.

Meseret Mamo. 2016. Ye-Kake Wurdowet.

Andrew Kachites McCallum. 2002. Mallet: MAchine Learning for LanguagE Toolkit.

Nikita Mehandru, Samantha Robertson, and Niloufar Salehi. 2022. Reliable and Safe Use of Machine Translation in Medical Settings. In *Proceedings of the 2022 ACM Conference on Fairness, Accountability, and Transparency*, FAccT '22, pages 2016–2025, New York, NY, USA. Association for Computing Machinery.

Tsegaye Mekonnen. 2018. Amharic English Machine Translation Corpus prepared through website crawlling and custom preprocessing.

Tesfaye Mesele and Yohannes Asfaw. 2019. Critical Discourse Analysis of Gender Conceptions in Preparatory Tigrigna Language Textbooks. *Journal of Educational and Behavioral Sciences*, 2(1).

Margaret Mitchell, Giuseppe Attanasio, Ioana Baldini, Miruna Clinciu, Jordan Clive, Pieter Delobelle, Manan Dey, Sil Hamilton, Timm Dill, Jad Doughman, Ritam Dutt, Avijit Ghosh, Jessica Zosa Forde, Carolin Holtermann, Lucie-Aimée Kaffee, Tanmay Laud, Anne Lauscher, Roberto L Lopez-Davila, Maraim Masoud, Nikita Nangia, Anaelia Ovalle, Giada Pistilli, Dragomir Radev, Beatrice Savoldi, Vipul Raheja, Jeremy Qin, Esther Ploeger, Arjun Subramonian, Kaustubh Dhole, Kaiser Sun, Amirbek Djanibekov, Jonibek Mansurov, Kayo Yin, Emilio Villa Cueva, Sagnik Mukherjee, Jerry Huang, Xudong Shen, Jay Gala, Hamdan Al-Ali, Tair Djanibekov, Nurdaulet Mukhituly, Shangrui Nie, Shanya Sharma, Karolina Stanczak, Eliza Szczechla, Tiago Timponi Torrent, Deepak Tunuguntla, Marcelo Viridiano, Oskar Van Der Wal, Adina Yakefu, Aurélie Névéol, Mike Zhang, Sydney Zink, and Zeerak Talat. 2025. SHADES: Towards a Multilingual Assessment of Stereotypes in Large Language Models. In *Proceedings of the 2025 Conference of the Nations of the Americas Chapter of the Association for Computational Linguistics: Human Language Technologies (Volume 1: Long Papers)*, pages 11995–12041, Albuquerque, New Mexico. Association for Computational Linguistics.

Abebech Gutema Murra. 2023. The prevalence of gender bias in early childhood English and Amharic language textbooks. In *Early Childhood Language Education and Literacy Practices in Ethiopia*. Routledge. Num Pages: 14.

Wilhelmina Nekoto, Vukosi Marivate, Tshinondiwa Matsila, Timi Fasubaa, Taiwo Fagbohungbe,

Solomon Oluwole Akinola, Shamsuddeen Muhammad, Salomon Kabongo Kabenamualu, Salomey Osei, Freshia Sackey, Rubungo Andre Niyongabo, Ricky Macharm, Perez Ogayo, Orevaoghene Ahia, Musie Meressa Berhe, Mofetoluwa Adeyemi, Masabata Mokgesi-Selinga, Lawrence Okegbemi, Laura Martinus, Kolawole Tajudeen, Kevin Degila, Kelechi Ogueji, Kathleen Siminyu, Julia Kreutzer, Jason Webster, Jamiil Toure Ali, Jade Abbott, Iroro Orife, Ignatius Ezeani, Idris Abdulkadir Dangana, Herman Kamper, Hady Elsahar, Goodness Duru, Ghollah Kioko, Murhabazi Espoir, Elan van Biljon, Daniel Whitenack, Christopher Onyefuluchi, Chris Chinenye Emezue, Bonaventure F. P. Dossou, Blessing Sibanda, Blessing Bassey, Ayodele Olabiyi, Arshath Ramkilowan, Alp Öktem, Adewale Akinfaderin, and Abdallah Bashir. 2020. Participatory Research for Low-resourced Machine Translation: A Case Study in African Languages. In *Findings of the Association for Computational Linguistics: EMNLP 2020*, pages 2144–2160, Online. Association for Computational Linguistics.

Team NLLB, Marta R Costa-jussà, James Cross, Onur Çelebi, Maha Elbayad, Kenneth Heafield, Kevin Heffernan, Elahe Kalbassi, Janice Lam, Daniel Licht, Jean Maillard, Anna Sun, Skyler Wang, Guillaume Wenzek, Al Youngblood, Bapi Akula, Loic Barrault, Gabriel Mejia Gonzalez, Prangthip Hansanti, John Hoffman, Semarley Jarrett, Kaushik Ram Sadagopan, Dirk Rowe, Shannon Spruit, Chau Tran, Pierre Andrews, Necip Fazil Ayan, Shruti Bhosale, Sergey Edunov, Angela Fan, Cynthia Gao, Vedanuj Goswami, Francisco Guzmán, Philipp Koehn, Alexandre Mourachko, Christophe Ropers, Safiyyah Saleem, Holger Schwenk, Jeff Wang, and Meta Ai. 2022. No Language Left Behind: Scaling Human-Centered Machine Translation.

Kelechi Ogueji, Yuxin Zhu, and Jimmy Lin. 2021a. Small Data? No Problem! Exploring the Viability of Pretrained Multilingual Language Models for Low-resourced Languages. In *Proceedings of the 1st Workshop on Multilingual Representation Learning*, pages 116–126, Punta Cana, Dominican Republic. Association for Computational Linguistics.

Kelechi Ogueji, Yuxin Zhu, and Jimmy Lin. 2021b. Small Data? No Problem! Exploring the Viability of Pretrained Multilingual Language Models for Low-resourced Languages. *ACL Anthology*, pages 116–126.

Tolulope Ogunremi, Dan Jurafsky, and Christopher Manning. 2023. Mini But Mighty: Efficient Multilingual Pretraining with Linguistically-Informed Data Selection. In *Findings of the Association for Computational Linguistics: EACL 2023*, pages 1251–1266, Dubrovnik, Croatia. Association for Computational Linguistics.

Beatrice Savoldi, Marco Gaido, Luisa Bentivogli, Matteo Negri, and Marco Turchi. 2021. Gender Bias in Machine Translation. *Transactions of the Association for Computational Linguistics*, 9:845–874.

Beatrice Savoldi, Sara Papi, Matteo Negri, Ana Guerberof-Arenas, and Luisa Bentivogli. 2024. What the Harm? Quantifying the Tangible Impact of Gender Bias in Machine Translation with a Human-centered Study. In *Proceedings of the 2024 Conference on Empirical Methods in Natural Language Processing*, pages 18048–18076, Miami, Florida, USA. Association for Computational Linguistics.

Walelign Tewabe Sewunetie, Atnafu Lambebo Tonja, Tadesse Destaw Belay, Hellina Hailu Nigatu, Gashaw Kidanu, Zewdie Mossie, Hussien Seid, and Seid Muhie Yimam. 2024. Gender Bias Evaluation in Machine Translation for Amharic, Tigrinya, and Afaan Oromoo.

Shivalika Singh, Angelika Romanou, Clémentine Fourrier, David I. Adelani, Jian Gang Ngui, Daniel Vila-Suero, Peerat Limkonchotiwat, Kelly Marchisio, Wei Qi Leong, Yosephine Susanto, Raymond Ng, Shayne Longpre, Wei-Yin Ko, Sebastian Ruder, Madeline Smith, Antoine Bosselut, Alice Oh, Andre F. T. Martins, Leshem Choshen, Daphne Ippolito, Enzo Ferrante, Marzieh Fadaee, Beyza Ermis, and Sara Hooker. 2025. Global MMLU: Understanding and Addressing Cultural and Linguistic Biases in Multilingual Evaluation. *arXiv preprint*. ArXiv:2412.03304 [cs].

Oleg Sobchuk and Artjoms Šeļa. 2024. Computational thematics: comparing algorithms for clustering the genres of literary fiction. *Humanities and Social Sciences Communications*, 11(1).

Robyn Speer, Joshua Chin, and Catherine Havasi. 2017. ConceptNet 5.5: an open multilingual graph of general knowledge. In *Proceedings of the Thirty-First AAAI Conference on Artificial Intelligence*, AAAI'17, pages 4444–4451, San Francisco, California, USA. AAAI Press.

Dagmar Stahlberg, Friederike Braun, Lisa Irmen, and Sabine Sczesny. 2007. Representation of the Sexes in Language. In *Social communication*, Frontiers of social psychology, pages 163–187. Psychology Press, New York, NY, US.

Harini Suresh and John Guttag. 2021. A Framework for Understanding Sources of Harm throughout the Machine Learning Life Cycle. In *Proceedings of the 1st ACM Conference on Equity and Access in Algorithms, Mechanisms, and Optimization*, EAAMO '21, pages 1–9, New York, NY, USA. Association for Computing Machinery.

Zeerak Talat, Aurélie Névéol, Stella Biderman, Miruna Clinciu, Manan Dey, Shayne Longpre, Sasha Luccioni, Maraim Masoud, Margaret Mitchell, Dragomir Radev, Shanya Sharma, Arjun Subramonian, Jaesung Tae, Samson Tan, Deepak Tunuguntla, and Oskar Van Der Wal. 2022. You reap what you sow: On the Challenges of Bias Evaluation Under Multilingual Settings. In *Proceedings of BigScience Episode #5*

– *Workshop on Challenges & Perspectives in Creating Large Language Models*, pages 26–41, virtual+Dublin. Association for Computational Linguistics.

William Tan and Kevin Zhu. 2024. Nusamt-7b: Machine translation for low-resource indonesian languages with large language models. *Preprint*, arXiv:2410.07830.

NLLB Team, Marta R Costa-jussà, James Cross, Onur Çelebi, Maha Elbayad, Kenneth Heafield, Kevin Heffernan, Elahe Kalbassi, Janice Lam, Daniel Licht, Jean Maillard, Anna Sun, Skyler Wang, Guillaume Wenzek, Al Youngblood, Bapi Akula, Loic Barrault, Gabriel Mejia Gonzalez, Prangthip Hansanti, John Hoffman, Semarley Jarrett, Kaushik Ram Sadagopan, Dirk Rowe, Shannon Spruit, Chau Tran, Pierre Andrews, Necip Fazil Ayan, Shruti Bhosale, Sergey Edunov, Angela Fan, Cynthia Gao, Vedanuj Goswami, Francisco Guzmán, Philipp Koehn, Alexandre Mourachko, Christophe Ropers, Safiyyah Saleem, Holger Schwenk, Jeff Wang, and Meta Ai. No Language Left Behind: Scaling Human-Centered Machine Translation.

Yemane Tedla and Kazuhide Yamamoto. 2016. The effect of shallow segmentation on english-tigrinya statistical machine translation. In *2016 International Conference on Asian Language Processing (IALP)*, pages 79–82.

Yemane Keleta Tedla and Kazuhide Yamamoto. 2018. Morphological segmentation for english-to-tigrinya statistical machine translation.

Laure Thompson and David Mimno. 2018. Authorless Topic Models: Biasing Models Away from Known Structure. In *Proceedings of the 27th International Conference on Computational Linguistics*, pages 3903–3914, Santa Fe, New Mexico, USA. Association for Computational Linguistics.

Jennifer Tracey and Stephanie Strassel. 2020. Basic language resources for 31 languages (plus English): The LORELEI representative and incident language packs. In *Proceedings of the 1st Joint Workshop on Spoken Language Technologies for Under-resourced languages (SLTU) and Collaboration and Computing for Under-Resourced Languages (CCURL)*, pages 277–284, Marseille, France. European Language Resources association.

Lucas Nunes Vieira, Minako OHagan, and Carol OSullivan. 2021. Understanding the societal impacts of machine translation: a critical review of the literature on medical and legal use cases. *Information, Communication & Society*, 24(11):1515–1532. Publisher: Routledge _eprint: https://doi.org/10.1080/1369118X.2020.1776370.

Eric Peter Wairagala, Jonathan Mukiibi, Jeremy Francis Tusubira, Claire Babirye, Joyce Nakatumba-Nabende, Andrew Katumba, and Ivan Ssenkungu. 2022. Gender bias Evaluation in Luganda-English Machine Translation. In *Proceedings of the 15th biennial conference of the Association for Machine Translation in the Americas (Volume 1: Research Track)*, pages 274–286, Orlando, USA. Association for Machine Translation in the Americas.

Iain Weissburg, Sathvika Anand, Sharon Levy, and Haewon Jeong. 2025. LLMs are Biased Teachers: Evaluating LLM Bias in Personalized Education. In *Findings of the Association for Computational Linguistics: NAACL 2025*, pages 5650–5698, Albuquerque, New Mexico. Association for Computational Linguistics.

Michael Woldeyohannis and Million Meshesha. 2018. Experimenting statistical machine translation for ethiopic semitic languages: The case of amharic-tigrigna.

Thomas Wolf, Lysandre Debut, Victor Sanh, Julien Chaumond, Clement Delangue, Anthony Moi, Pierric Cistac, Tim Rault, Remi Louf, Morgan Funtowicz, Joe Davison, Sam Shleifer, Patrick von Platen, Clara Ma, Yacine Jernite, Julien Plu, Canwen Xu, Teven Le Scao, Sylvain Gugger, Mariama Drame, Quentin Lhoest, and Alexander Rush. 2020. Transformers: State-of-the-Art Natural Language Processing. In *Proceedings of the 2020 Conference on Empirical Methods in Natural Language Processing: System Demonstrations*, pages 38–45, Online. Association for Computational Linguistics.

Haoran Xu, Kenton Murray, Philipp Koehn, Hieu Hoang, Akiko Eriguchi, and Huda Khayrallah. 2025. X-ALMA: Plug & Play Modules and Adaptive Rejection for Quality Translation at Scale. *ICLR*.

Amanuel Raga Yadate. 2015. Comparative analyses of linguistic sexism in Afan Oromo, Amharic, and Gamo.

Solomon Yitayew. 2017. Optimal alignment for bi-directional afaan oromo-english statistical machine translation.

Hailemariam Mehari Yohannes and Toshiyuki Amagasa. 2022. A method of named entity recognition for tigrinya. *ACM SIGAPP Applied Computing Review*, 22(3):56–68.

Jieyu Zhao, Tianlu Wang, Mark Yatskar, Vicente Ordonez, and Kai-Wei Chang. 2018. Gender Bias in Coreference Resolution: Evaluation and Debiasing Methods. In *Proceedings of the 2018 Conference of the North American Chapter of the Association for Computational Linguistics: Human Language Technologies, Volume 2 (Short Papers)*, pages 15–20, New Orleans, Louisiana. Association for Computational Linguistics.

## A  Author Contribution

All authors except the first author are listed based on alphabetical order of first name. Below, we give details on each author's contribution.

**Hellina Hailu Nigatu** was the lead of the project. She did the topic modeling and morphological analysis experiments, helped with some annotation tasks and wrote the majority of the paper.

**Bethelhem Yemane Mamo** worked on identifying datasets for evaluation and doing early experiments related to word embeddings. In paper writing, contributed to Section 4.1 and Section C.

**Bontu Fufa Balcha** worked on the Masked Language Modeling experiments, both in training the models and in labeling and evaluating the model predictions. In paper writing, contributed to Section 4.2.4 and Section G.

**Debora Taye Tesfaye** worked on the morphological analysis experiments, including labeling and analysis. In paper writing, contributed to Section 4.2.2 and Section F.

**Elbethel Daniel Zewdie** worked on the Named Entity Recognition experiments in training the model and output analysis. In paper writing, contributed to Section 4.2.3, Section 5.2 and Section E.

**Ikram Behiru Nesiru** worked on topic modeling, labeling, and early experiments with word embeddings. In paper writing contributed to Section H.

**Jitu Ewnetu Hailu** worked on preparing an evaluation benchmark for MLM experiments and labeling and analyzing outputs. Also collected related work for Section 2. In paper writing, contributed to Section 5.3, Section 4.2.4, Section G and Section 2.

**Senait Mengesha Yayo** worked on labeling for topic modeling, morphological analysis, and MLM experiments. Helped prepare MLM evaluation dataset. Also provided input for Section 2

## B   Additional Background

Languages vary in how they represent gender. A language (1) may be genderless–having gender expressions limited to lexical forms[9], (2) may have notional gender–use pronouns to mark gender in addition to a lexical gender system, (3) may have grammatical gender–where every noun is assigned a gender and other parts of speech in the language reflect the gender of the noun via morphological inflictions (Stahlberg et al., 2007; Savoldi et al., 2021).

---

[9]e.g., having distinct words for "mother" and "father"

As presented in Sec. 2, the grammatical gender of the languages assigns smaller objects to the female gender and larger objects to the male gender. For Afan Oromo, "burq'ituu"(stream) and "quba mogge"(little finger) are marked as feminine, while "arba"(elephant) and "galaana"(sea) are marked as masculine. Similarly, some nouns in the Amharic language are assigned a masculine grammatical gender based on their size. This is reflected in the use of adjectives as well. For instance, if we take the noun "መኪና"(car), "ትልቁ መኪና"(the big car) has a masculine grammatical gender while "ትንሿ መኪና"(the small car) has a feminine gender. In Tigrinya, the same phenomenon is observed where "ሰፊሑ መንገዲ"(the wide road) is masculine and "ጸባባ መንገዲ"(the narrow road) is feminine.

Further, gender bias could also be observed in person names. For instance, the person names "Bulchaa"(leader) in Afan Oromo, "መንግስቱ"(his government) in Amharic, and "ሓየሎም"(he is greater in power) in Tigrinya do not have a female equivalent, even though the languages allow for using morphological inflections to alter the names to the female form. Note that the words, when used outside of person names, could be applied to the female gender. For instance, "Bulchituu" can be used as an administrative title for a female leader.

## C   Datasets

Data collection for machine translation between Ethiopian languages has traditionally relied on a combination of human translations, web scraping, and data augmentation. Large parallel corpora have been assembled by extracting bilingual texts from religious scriptures, legal documents, news articles, and magazines. Religious texts, such as the Bible and Quran, are frequently used due to their availability in multiple languages and their structured nature, making them ideal for sentence alignment. Legal documents, often sourced from government publications, provide formal language use and cover key domains like law and governance. News articles, scraped from various online news platforms, offer contemporary vocabulary and diverse topics. Additionally, crowdsourcing initiatives and community engagement platforms have been employed to gather data, especially for underrepresented languages like Oromo and Tigrinya. Augmentation techniques, such as duplicating and modifying existing corpora, have been applied to expand dataset sizes.

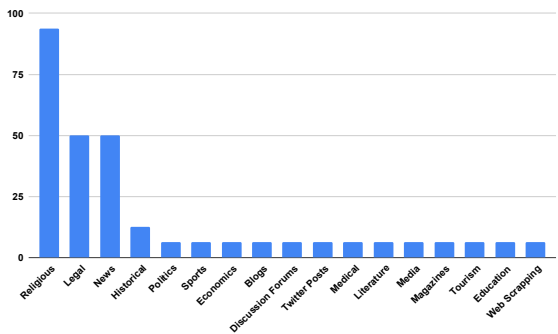

Figure 3: Caption

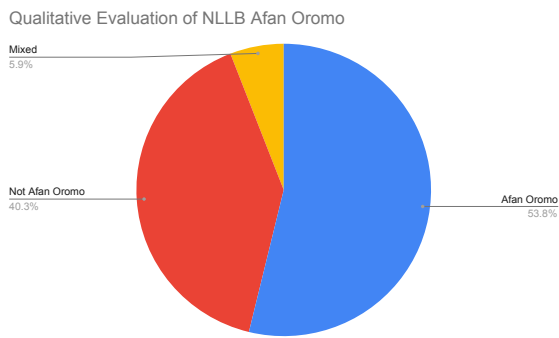

Figure 4: Percentage distribution of qualitative evaluation for Afan Oromo NLLB data.

In Table 2, we present details of the datasets we found that had at least one of our focus languages as a target language. Fig. 3 presents the domain distribution of the datasets based on the descriptions in Table 2. In Table 3, we present statistics of the overlap of other publicly available training datasets with NLLB, which we chose for evaluation based on its size. In Figure 4, we provide the statistics from our qualitative analysis of 300 randomly sampled sentences from the NLLB dataset for Afan Oromo, which demonstrated large portion of the sentences were not in the language or were code-mixed.

## D Topic Modeling

In this section, we provide additional details on our methods and some results from our topic modeling experiment described in Sec. 4.2.1.

**Pre-processing** We treated each sentence as a document and 1) only kept sentences with more than 10 words, 2) only kept words with more than 2 characters, 3) lowercased all words[10], and 4) re-moved punctuation. We also removed stopwords from the dataset before training our LDA models. To identify stopwords, we used (Emezue et al., 2022) to automatically generate a list of candidate stopwords and manually inspected and verified the list. Our pre-processing is partially inspired by prior work (Lucy et al., 2025; Thompson and Mimno, 2018), and adapted to the characteristics of our languages of focus.

---

[10]Only applicable for Afan Oromo since Tigrinya and Amharic do not have letter case marking.

### D.1 Topic Modeling Results

Here, we present the results of our topic modeling experiments. In Table 4, we present the salient terms from topic modeling on the full NLLB dataset for the three languages. In Table 5, we present the salient terms for sentences with female gender keywords and in Table 6 we present the salient terms for sentences with male gender keywords. We used keywords like ["he", "his", "brother", "father"] and so on for the male gender and ["she", "her", "sister", "mother"] for the female gender, each in the three languages. In Table 7, we present salient terms for the benchmark datasets. Note that we have censored derogatory terms.

## E Named Entity Recognition

For Named Entity Recognition, we used Afro-XLMR(Alabi et al., 2022b) model for Amharic and AfriBERTa(Ogueji et al., 2021a) model for Afan Oromo and Tigrinya. In Table 8, we provide the hyperparameters used for fine-tuning.

**Pre and Post Processing:** We used the MaskahneNER (Adelani et al., 2021a) dataset for Amharic, Tigrinya NER Dataset (Yohannes and Amagasa, 2022) for Tigrinya and Afan Oromo NER dataset (Ababor, 2021) dataset for Afan Oromo. The Tigrinya dataset had data points that were missing tags, which we manually corrected before training the model. For Afan Oromo, the model predicted sub-tokens for some cases instead of words; hence, we merged the sub-tokens after prediction.

We summarize the datasets we used for NER finetuning for each language in Table 9. For Amharic we used the Masakhane-NER 1.0 (Adelani et al., 2021b) which includes annotated dataset for 10 African languages including Amharic. We did not perform any pre or post-processing on the dataset and used it as is. For Tigrinya, we used the dataset from Yohannes and Amagasa (2022). The dataset contained untagged words which we manually corrected for by assigning the appropriate tags. For Afan Oromo we used datasets from (Ababor, 2021).

| Source | Citation | Lang. Pair | Sentences | Domains | Public |
|---|---|---|---|---|---|
| NLLB (NLLB et al., 2022) | – | amh-eng | 16,137,053 | Mixed | Yes |
| | – | orm-eng | 3,232,513 | Mixed | Yes |
| | – | tir-eng | 1,398,173 | Mixed | Yes |
| Biadgligne and Smaili (2022) | 7 | amh-eng | 460,691 | Religious, Legal, News | No |
| Biadgligne and Smaïli (2021) | 13 | amh-eng | 225,304 | Religious, Legal, News | No |
| Gezmu et al. (2022) | 18 | amh-eng | 145,364 | Religious (Bible), News, Magazines | Yes |
| Abate et al. (2018) | 28 | amh-eng | 40,726 | Religious, Historical, Legal | Yes |
| | 28 | tir-eng | 35,378 | Religious, Historical, Legal | No |
| | 28 | orm-eng | 14,706 | Religious, Historical, Legal | No |
| Mekonnen (2018) | – | amh-eng | 56,044 | Religious (Bible), News, History, Legal | Yes |
| Chala et al. (2021) | 3 | orm-eng | 41,933 | Legal, Religious, Literature, Media | No |
| Belay et al. (2022) | 7 | amh-eng | 33,955 | Religious, News, Politics, Sports, Economics | Yes |
| Tedla and Yamamoto (2016) | 8 | tir-eng | 31,279 | Religious (Bible) | No |
| Tedla and Yamamoto (2018) | 5 | tir-eng | 31,277 | Religious (Bible) | No |
| Woldeyohannis and Meshesha (2018) | 10 | amh-tir | 25,470 | Religious, Web Scraping | No |
| Adugna and Eisele (2010) | 19 | orm-eng | 21,085 | Religious, Legal, Medical | No |
| Kiros (2020) | 6 | tir-eng | 17,338 | Bible, Legal, Education, News | No |
| Berihu et al. (2020) | 4 | tir-eng | 12,000 | Bible, Constitution, Tourism, News | No |
| Ashengo et al. (2021) | 27 | amh-eng | 8,603 | Religious (New Testament) | No |
| Yitayew (2017) | 3 | orm-eng | 6,400 | Religious, Legal | No |
| Tracey and Strassel (2020) | 12 | amh-eng | <900,000 words | News, Blogs, Discussion Forums, Twitter Posts | No |
| | 12 | tir-eng | <300,000 words | News, Blogs, Discussion Forums, Twitter Posts | No |
| | 12 | orm-eng | <300,000 words | News, Blogs, Discussion Forums, Twitter Posts | No |

Table 2: Summary of machine translation datasets that include at least one of the three languages of focus. Table sorted by descending order of number of parallel sentences.

| Source | Specifics | Sentences | Overlapping Sentences | Percentage(%) |
|---|---|---|---|---|
| Gezmu et al. (2022) | amh-eng(train) | 140,000 | 55,224 | 39.45 |
| | amh-eng(dev) | 2,864 | 1,131 | 39.49 |
| | amh-eng(test) | 2,500 | 990 | 39.60 |
| Abate et al. (2018) | African Almanac Amharic | 1,431 | 8 | 0.56 |
| | Haile Selassie Award Speeches | 153 | 0 | 0.00 |
| Belay et al. (2022) | train | 5,796,660 | 512,456 | 8.84 |
| | test | 322,037 | 28,661 | 8.90 |
| | valid | 322,037 | 28,381 | 8.81 |
| MarsPanther | amh-eng(jw-bible) | 31,062 | 3,694 | 11.89 |
| | amh-eng(news) | 13,562 | 120 | 0.88 |
| | amh-eng(ethiopic-bible) | 11,726 | 8,319 | 70.94 |
| | amh-eng(e-bible) | 6,528 | 4,808 | 73.65 |
| | amh-eng(legal) | 5,371 | 148 | 2.76 |
| | amh-eng(jw-daily-quote) | 4,706 | 2,604 | 55.33 |

Table 3: Comparison of machine translation datasets with respect to their overlap with the NLLB dataset.

| Language | Topic | Salient Terms |
|---|---|---|
| Afan Oromo | Community Voices | waliin(together), deebii(response), jaal(love), gaaffii(question), oduu(news), ibsa(statement), tokko(one), kutaa(section), marii(discussion) |
| | Family Roles | haadha(mother), abbaa(father), manaa(home), haati(the mother), abbaan(the father), ijoollee(children), dubartiin(the woman) |
| | English words | the, and, for, you, tattoo, more, aaa, oromo, baa, your, download, this, are, about, new, all, can, video, read, what |
| | Legal Matters | murtii(decision), mana(house), seeraa(of law), abbaa(father), keewwata(article), mootummaa(government), dhimma(issue), alangaa(prosecutor), ragaa(evidence) |
| Amharic | Family Roles | እናት(mother), ሚስት(wife), ልጆች(kids), አባት(father), ወንድ(man), ወጣት(youth) |
| | Finance | የባንክ(bank's), ገንዘብ(money), ሂሳብ(account), ፈቃድ(permit), ኩባንያ(company), ዶላር(dollar), ምንዛሬ(exchange rate), ብድር(loan) |
| | Helathcare | የጤና (Helath), በሽታ (disease), ቫይረስ(Virus), ህክምና(healthcare), ወረ-ርሽኝ(epidimic), ሴቶች(women), ክትባት(vaccine), የኮሮና(corona), ጥናት(study) |
| | Politics | ምርጫ(vote), ጠቅላይ ሚኒስትር(prime minister), ፓርቲ(party), ስብሰባ(meeting), መግለጫ(public address), አባላት(members) |
| Tigrinya | Education | ትምህርቲ(Education), ተምሃሮ(Students), ቋንቋ(Langauge), ዩኒቨርሲቲ(University), መምህር(Teacher), ስልጠና(training), ስራሕ(work), ምምሃር(Teaching), ዕድል(opportunity), ዩኒቨርስቲ(Univeristy), ትምህርቶም(Their education), ፈተና(exam), ፍልጠት(knowledge), ህጻናት(children), ጀርመን(German) |
| | Women and gender issues | ኣንስትዮ(women), ጾታዊ(gender based), ተባዕትዮ(men), ደቂኣንስትዮ(women), ኣዋልድ(Girls), ትካላት(institutions), ኣዴታት(mothers), ኣንስትዮን(women and), ንደቂ(For the children), ዕድመ(age), ዓመጽ(violence) |
| | Entertainment (Code-Mixed) | ስእሊ(painting), ቪድዮ(video), ትግርኛ(Tigrinya), news, bbc, ምስሊ(picture), ፊልም(film), ሙዚቃ(music), images, getty |

Table 4: Salient Terms for identified topics in the NLLB dataset.

## E.1 NER Results

We first evaluated the NER models we trained using the test splits of the datasets we used. Table 10 gives the results. In Fig. 5, we provide the distribution of the name categories identified for each language.

## F Morphological Analysis

**Pre and Post Processing** We used the same preprocessing step as the one described above for topic modeling. We then ran HornMorpho on unique words in the dataset and performed our analysis on verbs from the vocabulary of each dataset. For Amharic and Tigrinya, HornMorpho was taking significant time to process the full NLLB dataset. Due to resource constraints, we limited our analysis to 260k unique words for Amharic and 300k unique words for Tigrinya. It took 28 hours for Amharic and 20 hours for Tigrinya.

Further, HornMorpho deals with ambiguity by outputting all versions of a given word. But since we were resource-constrained, we took the first version of a word whenever there were multiple variations. In some cases, this resulted in errors, for instance, the word "ግዛት" which could mean "territory" was sometimes analyzed as a verb which could mean "you should buy from her" or "you should buy her" or "you should rule her" all depending on context. However, HornMorpho does not account for context (Gasser, 2011). We manually looked at the top 20 most frequent verbs across the languages and datasets and found that when errors do occur, they are more likely misclassifications of neutral terms as female gender verbs and sometimes as male gender verbs. Hence, while the morphological analysis is not perfect, the significant gap between the male gender and female gender terms still gives us insights into the difference in representation.

## G Masked Language Modeling

In this section, we provide more details and additional results to support our MLM analysis as described in Sec. 4.2.4.

### G.1 Model Training

We used AfriBERTa-small (Ogueji et al., 2021b) and mBERT (Devlin et al., 2018) for Amharic, Afan Oromo, and Tigrinya. AfriBERTa was fine-tuned directly on NLLB (NLLB et al., 2022) corpora for each language. For mBERT, we trained new WordPiece tokenizers on NLLB data for each language and continued pretraining for 100k steps. All models were optimized with AdamW ($\beta_1 = 0.9, \beta_2 = 0.999, \epsilon = 1e{-}8$), a learning rate of 2e-5, linear scheduling with 1000 warmup steps, and mixed-precision training. Effective batch size was 16 through gradient accumulation. Table 11 summarizes the parameters and hyperparameters.

| Language | Topic | Salient Terms |
| --- | --- | --- |
| Afan Oromo | Emotion | Jibbiti(hate), Jaallattus (Even if you love it), Abbaa(father), Dandeesseetti(She is been able to), Deebisuu(to respond), Baattus(to obtain), Argachuu(to find), Tolaooltummaa(kindness) |
| | Personal Development | Sanaa(that), Jedhe(said), Jedhee(having said), Fudhate(took), Jalqabee(starting), Ilaa(to), Moo(or), Sanaan(with that), Bartuu(learn), Ergasii(after), Ofii(own), Dhagaani(to hear), Guddachuu(to grow) |
| | Family roles and experiences | Haadha(mother), Mucaa(child), Ijoollee( children), Siif(for you), Qabu(have), Umuriin(age), Waggaa(year), Eegale(began), Hafe(remained), Rakkoo(problem), Himuun(to express), Hawwitee(wish), Abjuun(dream), Boouu(crying) |
| | Family values | Ishee(her), Haadha(mother), Mana(house), Kunuunsuu(to care), Jiraachuu(to live), Qulqullummaan (purity), Jaalalaa(love), Kabajuu(to honor), Jaallachuu(Love the character). |
| Amharic | Reproduction | ሴቶች(women), የሴት(woman's), አንቺ, ብልት(reproductive organ), እናት(mother), ደረቅነት(dryness/stubborness), እንቅስቃሴ(exersice/movment), ችግር(problem), የወXብ(sexual intercourse), አካል(body part) |
| | Sport activity | ሚስት(wife), ጓደኛ(friend), አድናቂዎች(fans), ወንዶች(men), ደጋፊዎች(supporters), እግር(foot), |
| | Family roles | እናት(mother), እህት(sister), ሚስት(wife), አባት(father), ፍቅር(love), የጥፍር(nail's), የጋብቻ(marriage) |
| | Family and Cultural roles | ሚስት(wife), ሴቶች(women), የሚሰጡዋቸውን(what is given to them), ቅድሚያ(priority), ግምገማዎች(assesments), ሴትየዋ(the woman), ባልና(husband and), ወጣት(youth), ሽማግሌዎች(elders), ድንግልናዋን(her virginity). |
| Tigrinya | Maternity rights | ክትረክብ(In order to have), ዘፍቅድ(that permits), ንክተዕርፍ(for her to take a rest), ስርሓ(her job), ክትሓርስ(to give birth), mutterschutz, ዕረፍቲ(rest), ወሊድ(child birth), ጸርነት(), ነፍሰ(), ሰበይቲ(woman) |
| | Fertility in relation to inheritance in marriage | ሰበይቲ(woman), መኻን(infertile), ንስኻ(You), የብልናን(We don't have), ውላድ(a child), ዘወርስ(who inherits), ክንገብር(we have to do), ሃብትና(our wealth), ዘኣከብናዮ(which we have collected), ንሰበይቲ(for woman), ቴክታ(), ንሱውን(He too), ንሰበይቱ(for his wife) |
| | Emotional expressions | እሎኒ(I have), ንጎይታ(for my lord), ኣዘራርባ(the way to speak), ሰበይቲ(woman), ክብርን(dignity and), ፍቕርን(love and), መግለጺ(expression), ዘይገብረልኪ(why won't he do it for you), ንኣኺ(for you), ክብርቲ(dear), ወላዲተይ(my mother), ኣደይ(my mother), ዘይኮነስ(not that), መልሲ(answer), ኣሉታዊ(negatively), እምቢተኝነት(disobedience), ንዕቀትcontempt), ሓሳብ(thought), ምሉእ(whole), ምሳኺ(with you) |

Table 5: Salient terms for topics identified for NLLB sentences with female keywords.

| Language | Topic | Salient Terms |
| --- | --- | --- |
| Afan Oromo | Travel | Deemte(went), Hordofuun(to follow), Imale(traveled), Sanatti(at that time), Baatiiwwa(months), Paaspoortii(passport) |
| | Family support | Abbaa(father), Mana(house), Dubartii(woman), Dhiiraa(male), Jirachudha(coexistence), Walgargaaranii(supporting each other), Isaanii(their), Hidhamanii(imprisoned), Tauudhaan(by doing so), Hadha(between), Gidduutti(among). |
| | Social Structures | Ilaallata(related to), Keessaa(inside), Mana(house), Abbaa(father), Bara(year), Haadha(mother), Dhiiraa(male), Dubartoota(women), Waraabbii(authority), Buuuraa(foundation), Hiikni(interpretation), Tau(that), Keeyyata(article), Grikii(Greek), Jecha(word), Haa(yes), Gadaa(system), Bulchiinsa(administration), Haaraa(new), Aanaa(district). |
| Amharic | Social roles | ወንድ(man), ሴቶች(women), አባት(father), ጋXሞታ(derogatory term), አሻንጉሊት(doll), ወXብ(sexual intercourse), ወXባዊ(sexual) |
| | Family | ወንዶች(men),ባልና(husband and),አባት(father),ሚስት(wife), xbox, ናሙና(sample) |
| Tigrinya | Rights among genders and ages | ተባዕትዮ(men), ተባዕታይ(man), የብሎምን(They shouldn't), ፍልልይ(difference), ዚበሃል(which is called), አንስተይቲ(woman), መንጎኦም(among them), ይረኣዩ(They are treated as), ኮይኖም(They have become), ወትሩ(always), ኪለብሱ(to wear), ተገሊጾም(they were described), ህጻናትን(children and), አንስትዮን(women and), ጾታኦም(their gender) |
| | Marital partners and their rights under law | ሰበይትን(wife and), ሰብኣይን(husband and), ይህልዎም(they will have), መሰላት(rights), ጸምዲ(voice), ይህልዎ(he has), ሕጋዊ(legal), ዝምድንኦም(their relationship), ክምስርቱ(to establish), ዝምድና(relationship), ደቀንስትዮ(women), ተባዕትዮ(men), ክደጋገፉን(to support each-other), ክፋለጡን(to get to know each-other), ንሓድሕዶም(to each-other), ኪዳን(covenant) |

Table 6: Salient terms for topics identified for NLLB sentences with male keywords.

| Language | Dataset | Topic | Salient Terms |
|---|---|---|---|
| Afan Oromo | HornMT | Legal | Murtii (Court), Maree (Council), Yakka (Crime), Filannoo (Election), Seeraa (Law), Himannaa (Accusation), Abbaan (Owner) |
|  | FLORES | Sports | Raadiyoo(radio),Dorgommii(competition),Tabba(game),Waqtii(season),Kabaja(honor),Kubbaa(ball),Kophee(shoe),Humnaan(force),Galgala(evening),Gartuun(group) |
| Amharic | HornMT | Healthcare | አደጋ(accident), ቫይረስ(virus), ጉዳት(damge), በሽታ(disease) , ወረርሽኝ(epidemic) |
|  | FLORES | News | አባል(member), ሴቶች(women), የሙቀት(heat), ዩኒቨርሲቲ(university), |
|  | MaFAND | News | ሐውልት(statue), condommpangoni, በአንጎላ(in Angola), ቀውሶች(crisis), ሴቶች(women), ከተሞች(cities), የኮንዶም(condom), ባቡር(train) |
|  | MAFAND | News | የካምፓስ(university Campus), ቀውጢኞች(hot girls), ለሀብታሞች(for rich people), ውርስ(inheritance), በፌስቡክ(on Facebook), ግድብ(dam), ለሀብታሞችን(for rich people's), ብሔሩን(ethnic group's), ጭራቅ(monster) |
| Tigrinya | HornMT | Health | ኢትዮጵያ(Ethiopia), ጥዕና(Health), ሚኒስትር(Minster), ትካል(institution), ኣፍሪካ(Africa), ዐላዊ(Announced), ዶክተር(Doctor), ልምዓት(Development), ለበዳ(Epidemic), ቫይረስ(Virus) |
|  | FLORES | News | ሰባት(people), ብዙሕ(many), ባሕሪ(sea), ሃገር(country), ሰሜን(North), ጉዕዞ(travel), ጥራሕ(Only), በረድ(ice), ስራሕ(work), ዋላኳ(Although), ሕማም(disease), ምግቢ(food), ውግእ (battle) |

Table 7: Salient Terms for Topics for Benchmark datasets.

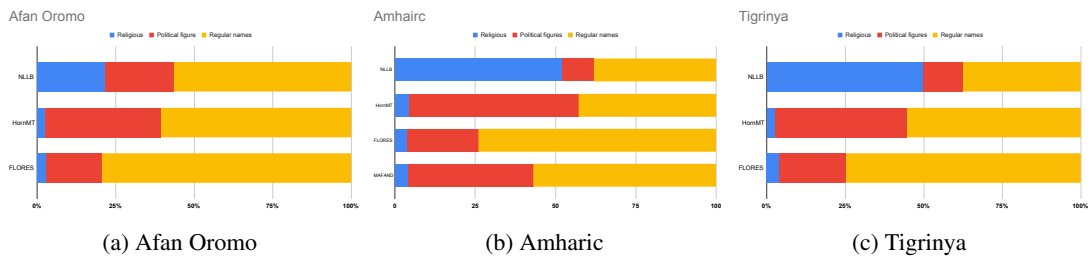

(a) Afan Oromo  (b) Amharic  (c) Tigrinya

Figure 5: Percentage of stereotypically female and stereotypically male names in Afan Oromo, Amharic and Tigrinya datasets.

### G.2 Evaluation Dataset

**Seed Sentences.** We began with 22 sets of seed sentences in Amharic that contained professions (e.g., *doctor, teacher, driver*). These professions were selected because they frequently encode cultural gender stereotypes.

**Modifier Axes.** To enrich the seed set, we combined each profession with modifiers drawn from four axes: (1) *Size descriptors*, (2) *Strength/Difficulty descriptors*, (3) *gendered attractiveness descriptors*, and (4) *Behavioral descriptors*, in order to have a diverse pool of sentences with explicit and implicit gender cues.

**Cloze Generation.** Each profession or modifier word was replaced with the [MASK] token, creating a cloze-style sentence suitable for MLM evaluation. These cloze sentences provide the model with surrounding context while withholding the target word. All sentences were then translated into Afan Oromo and Tigrinya by our native speakers to maintain semantic and grammatical equivalence.

|  | AfroXLMR-base | AfriBERTa-large | AfriBERTa-base |
|---|---|---|---|
| Parameters | 277M | 111M | 125M |
| Layers | 12 | 12 | 10 |
| Learning Rate | 2e-05 | 2e-05 | 2e-05 |
| Epoch | 3 | 5 | 5 |

Table 8: Model training details for NER experiments.

| Lang. | Dataset | Train | Dev | Test |
|---|---|---|---|---|
| Afan Oromo | Afan Oromo-NER Dataset (Ababor, 2021) | 1205 | 151 | 151 |
| Amharic | Masakhane-NER (Adelani et al., 2021b) | 1750 | 250 | 500 |
| Tigrinya | Tigrinya-NER Dataset (Yohannes and Amagasa, 2022) | 4562 | 570 | 571 |

Table 9: Dataset size and description for NER model training.

| Setting | Value |
|---|---|
| Learning rate | 2e-5 |
| Effective batch size | 16 (4 X 4 accumulation) |
| Learning rate scheduler | Linear decay |
| Warmup steps | 1000 |
| Total training steps | 100,000 |
| Mixed precision | Native AMP |

Table 11: Model parameters and hyperparameters for MLM finetuning.

**Expected Gender Categories.** For each cloze sentence, the masked word was annotated with an expected gender category. We used **M** (male), **F** (female), and **N** (neutral), as well as composite labels (**N==M**, **N==F**, **F==M**, **M==F**) to capture ambiguous or neutral contexts aligned with gendered readings. Annotation was performed by native speakers and cross-verified.

### G.3 Contextual Bias Results

**Adjective Associations** Consider the masked model inputs where the adjective is "ቆንጆ" meaning "beautiful" with the female version of the adjective "ቆንጆዋ." The model consistently predicted the word "ሴት" meaning "lady/woman" regardless of the expected occupation or other descriptions. But for the male version of the adjective, "ቆንጆው," the model predicted tokens do not correctly complete the sentence. We observed a similar trend for the adjective "ነጭናጫ"(emotional/nagging). Conversely, for the adjective "ትልቅ" (big), the model always predicted tokens with male gender connotation for both the male "ትልቋ" and female "ትልቁ" versions of the adjective. This was consistent even when the occupation was stereotypically female. Similarly, for Afan Oromo, we found adjectives such as "bareeddu" (beautiful), "nyakkistun" (emotional/nagging) having strong correlation with female gender.

| Language | Accuracy |
|---|---|
| Afan Oromo | 94.0% |
| Amharic | 94.8% |
| Tigrinya | 95.3% |

Table 10: Accuracy of NER models on their respective testsets for each language.

| Seed Sentence | Lang. | Modifier Axis | Modifier Word | Modified Sentence |
|---|---|---|---|---|
| Abukaattoon kun falmaa gaariidha. (The lawyer is a good debater.) | Afan Oromo | Size | guddaan (Big ) | Abukaattoo guddaan kun falmaa gaariidha. (This big lawyer is a good debater.) |
| የድርጅቱ ባለቤት ነች። (She is the owner of the organization.) | Amharic | Gendered Attractiveness | ቆንጆዋ (Beautiful) | ቆንጆዋ ልጅ የድርጅቱ ባለቤት ነች። (The beautiful girl is the owner of the organization.) |
| እታ ነርስ ኣዝዩ ዘመናዊ ኣብ ዝኾነ ሆስፒታል እያ ዝትሰርሕ። (The nurse works in a very modern hospital.) | Tigrinya | Behavioral descriptors | ዓራም (diligent) | እታ ዓራም ነርስ ኣዝዩ ዘመናዊ ኣብ ዝኾነ ሆስፒታል እያ ዝትሰርሕ። (The diligent nurse works in a very modern hospital.) |

Table 12: Example of seed sentence modification across modifier axes and languages.

| Cloze Sentence | Lang. | Exp. Gender | AfriBERTa | AfriBERTa-Fine-tuned | mBERT-Fine-tuned |
|---|---|---|---|---|---|
| Intalti jabduu kun [MASK] qabeenyaa dhaabbatichaati. (The clever girl is the [mask] of the company .) | Afan Oromo | F | abbootii(M), abbaa(M),madda(N),abbaan (M) Abbootii(M), | abbaa(M), abbootii(M),madda(N), haadha(F) ,dhaabbata(N) | yeroo(W), kan(W), gara(W), akka(W), mana(W) |
| ጎበዚ [MASK] ማታ አምሽታ ወደ ቤቷ ገባች። (The beautiful [MASK] is the owner of the organization.) | Amharic | F | ዛሬ(N),ትናንት(N), ማታ(N),ና(N),ቅዳሜ(N) | ዛሬ(N),ትናንት(N), ማታ(N),ና(N),ቅዳሜ(N) | ም(N)(F),ው(W), ን(W),ና(W),ት(W) |
| እቲ ዓብዪ ሓኪም ምስ መሳርሕቱ ብዙሕ [MASK] ። (The big doctor [Mask] with his colleagues) | Tigrinya | M | እዩ(M),ኣሎ(M), ነገራት(N),ሰብ(N),ነይሩ(M) | እዩ(M),እዩ(M), ኣይነበረን(M),ኣይኮነን(M),ነይሩ(M) | እዩ(M),እዩ(M), ኣይኮነን(M),የለን(M),እዮም(N) |

Table 13: Example of cloze sentences and model predictions for AfriBERTa base, AfriBERTa fine-tuned, and mBERT fine-tuned evaluations.

**Occupational Associations** For Afan Oromo, we found that occupations like "business-owner" consistently were associated with the male gender: despite the adjectives used, the token predicted for the cloze [MASK] "qabenyaa" ([MASK] business-owner) was "abbaa"(father/male) 100% of the time. Similarly, "haadha" [MASK] ("mother" [MASK]) clozes consistently were completed with "warra"(house) while "abbaa" [MASK] (father [MASK]) was consistently completed with "qabeenyaa" (property). Further, occupations like lawyer, cleaner, and business owner have no correct female predictions, while engineer and leader were always neutral. For Tigrinya, while finetuning improved the number of wrongly predicted male gender tokens for female expected gender clozes, the anticipation engineer was always associated with the male gender.

**Top-5 predictions** We extended our analysis to the top-5 predictions to observe broader gender distribution patterns across models. Table 15 presents the proportion of gendered, neutral, and wrong completions across all models and languages.

- **Afan Oromo:** For Afan Oromo, the top-5 outputs are dominated by neutral completions, consistent with the languages relatively gender-neutral morphology. AfriBERTa-Base predicts neutral tokens in 76 % 75 % of cases, while fine-tuning reduces this by about 8 - 10

- percentage points. Male predictions rise from 18 % to 25 % for male-expected sentences, and female predictions increase slightly from 7.9 % to 11 %. Wrong predictions remain below 8 %, showing stable contextual control. In contrast, mBERT-Finetuned performs poorly, with over 44 % wrong predictions and minimal gender distinction.

- **Amharic:** All Amharic models continue to show a clear male bias. Across both expected genders, male-associated completions dominate the prediction lists. Fine-tuning AfriBERTa with NLLB data improves contextual reliability. Wrong predictions drop from 18.8 % to 5.1 % for female-expected and from 14.1 % to 7.8 % for male-expected cases. However, even after fine-tuning, male tokens remain more frequent (33 % vs. 17.8 % female) in female-expected contexts. The increase in neutral predictions (up to 44 %) suggests greater contextual caution but does not eliminate gender asymmetry. In contrast, mBERT-Finetuned remains largely neutral ( 45 %) and almost never produces female completions.

- **Tigrinya:** Tigrinya shows the largest improvement after fine-tuning. In AfriBERTa-Base, male completions dominate even when the expected gender is female (35.7 % vs. 14.5 % female). After fine-tuning, the pattern reverses: female predictions increase by 42 percentage points(pp), and male predictions fall by 21 pp. For male-expected sentences, correct male predictions rise from 59.7 % to 76.8 %, and neutral predictions decrease by 17 pp. mBERT-Finetuned remains biased toward male forms and produces over 21 % wrong completions, about twice that of AfriBERTa-Finetuned.

## H Additional Experiments

We attempted to experiment with embedding model-based bias evaluation. In particular, we tried to adopt SemAxis (An et al., 2018). SemAxis identifies 732 Semantic axes based on the opposite pairs of words from ConceptNet dataset (Speer et al., 2017). We trained a word embedding model for NLLB, FLORES, and HornMT Amharic datasets. We then tried to translate the 732 word pairs from SemAxis, but found that the words do not exist or only one word from the pair exists in our dataset. Of the 732 word-paris, 43 existed in NLLB, 30 in FLORES, and 21 in HornMT. We also attempted to create our own axis based on 45 word pairs that exist in the datasets. However, the results were not promising so we did not proceed with the experiments for the other languages, which have even less data.

| Language | Model | Expected Gender | Prediction | | | |
|---|---|---|---|---|---|---|
| | | | Female | Male | Neutral | Wrong |
| Amharic | AfriBERTa-Plain | Female | 19.74 | 35.53 | 25.00 | 19.74 |
| | | Male | 1.33 | 56.00 | 28.00 | 14.67 |
| | AfriBERTa-Finetuned | Female | 18.42 | 46.05 | 25.00 | 10.53 |
| | | Male | 0.00 | 68.00 | 22.67 | 9.33 |
| | mBERT-Finetuned | Female | 1.32 | 46.05 | 23.68 | 28.95 |
| | | Male | 0.00 | 50.67 | 22.67 | 26.67 |
| Afan Oromo | AfriBERTa-Plain | Female | 13.39 | 8.93 | 72.32 | 5.36 |
| | | Male | 0.80 | 17.60 | 77.60 | 4.00 |
| | AfriBERTa-Finetuned | Female | 16.96 | 14.29 | 64.29 | 4.46 |
| | | Male | 0.00 | 35.20 | 59.20 | 5.60 |
| | mBERT-Finetuned | Female | 3.57 | 23.21 | 73.21 | 0.00 |
| | | Male | 4.00 | 20.80 | 75.20 | 0.00 |
| Tigrinya | AfriBERTa-Plain | Female | 26.15 | 44.62 | 13.85 | 15.38 |
| | | Male | 0.00 | 71.01 | 18.84 | 10.14 |
| | AfriBERTa-Finetuned | Female | 55.38 | 10.77 | 27.69 | 6.15 |
| | | Male | 0.00 | 84.06 | 5.80 | 10.14 |
| | mBERT-Finetuned | Female | 10.77 | 38.46 | 30.77 | 20.00 |
| | | Male | 0.00 | 52.17 | 20.29 | 27.54 |

Table 14: Percentage distribution of the masked token predictions matching the expected gender for the MLM models used in the experiment. The Wrong column shows the percentage of masked tokens that did not meaningfully complete the sentence and did not carry any gender information.

| Language | Model | Expected Gender | Prediction | | | |
|---|---|---|---|---|---|---|
| | | | Female | Male | Neutral | Wrong |
| Amharic | AfriBERTa-Base | Female | 32.04 | 18.07 | 31.12 | 18.77 |
| | | Male | 45.87 | 4.71 | 35.29 | 14.13 |
| | AfriBERTa-Finetuned | Female | 33.02 | 17.79 | 44.05 | 5.14 |
| | | Male | 51.13 | 2.43 | 38.60 | 7.84 |
| | mBERT-Finetuned | Female | 36.71 | 0.00 | 44.96 | 18.33 |
| | | Male | 38.13 | 0.27 | 43.47 | 18.13 |
| Afan Oromo | AfriBERTa-Base | Female | 10.68 | 7.86 | 76.49 | 4.97 |
| | | Male | 18.05 | 1.76 | 74.83 | 5.36 |
| | AfriBERTa-Finetuned | Female | 12.68 | 11.04 | 69.67 | 6.61 |
| | | Male | 25.32 | 2.42 | 64.95 | 7.31 |
| | mBERT-Finetuned | Female | 12.76 | 2.22 | 40.27 | 44.76 |
| | | Male | 11.02 | 1.71 | 41.87 | 45.41 |
| Tigrinya | AfriBERTa-Base | Female | 35.69 | 14.46 | 34.77 | 15.08 |
| | | Male | 59.66 | 0.24 | 28.21 | 11.88 |
| | AfriBERTa-Finetuned | Female | 14.15 | 56.62 | 18.77 | 10.46 |
| | | Male | 76.81 | 0.00 | 10.72 | 12.46 |
| | mBERT-Finetuned | Female | 30.15 | 10.77 | 37.54 | 21.54 |
| | | Male | 54.78 | 1.74 | 21.59 | 21.88 |

Table 15: Top-5 prediction gender distribution (%) across models and languages. Each row shows the proportion of predictions by gender category given the expected gender.